\documentclass[10pt]{article}

\usepackage[english]{babel}
\usepackage{amsmath}
\usepackage{amsfonts}
\usepackage{amssymb}
\usepackage{amsthm}
\usepackage{dsfont}
\usepackage{mathrsfs}
\usepackage{graphicx}
\usepackage{url}
\usepackage{caption}
\usepackage{subcaption}
\usepackage[section]{placeins}
\usepackage[round]{natbib}

\newcommand{\R}{\mathbb{R}}

\newcommand{\ud}{\,\mathrm{d}}
\def\P{\mathds{P}}

\DeclareMathOperator*{\argmax}{arg\,max}

\DeclareMathAlphabet{\mathpzc}{OT1}{pzc}{m}{it}

\usepackage{xcolor}

\allowdisplaybreaks

\begin{document}

\title{Modelling-based experiment retrieval:\\ A case study with gene expression clustering}

\author{Paul Blomstedt\thanks{Helsinki Institute for Information Technology HIIT, Department of Computer Science, Aalto University, Espoo, Finland}
	\and
    Ritabrata Dutta\footnotemark[1] $^1$ 
	\and
    Sohan Seth\footnotemark[1] $^2$
	\and
    Alvis Brazma\thanks{European Molecular Biology Laboratory, European Bioinformatics Institute, Wellcome Trust Genome Campus, Hinxton, UK} 
	\and
    Samuel Kaski\footnotemark[1] }

\date{}

\maketitle

\begin{abstract}

\textbf{Motivation:}
Public and private repositories of experimental data are growing to sizes that require dedicated methods for finding relevant data. To improve on the state of the art of keyword searches from annotations, methods for content-based retrieval have been proposed. In the context of gene expression experiments, most methods retrieve gene expression profiles, requiring each experiment to be expressed as a single profile, typically of case vs. control. A more general, recently suggested alternative is to retrieve experiments whose models are good for modelling the query dataset. However, for very noisy and high-dimensional query data, this retrieval criterion turns out to be very noisy as well. \\ 
\textbf{Results:}
We propose doing retrieval using a denoised model of the query dataset, instead of the original noisy dataset itself. To this end, we introduce a general probabilistic framework, where each experiment is modelled separately and the retrieval is done by finding related models. For retrieval of gene expression experiments, we use a probabilistic model called product partition model, which induces a clustering of genes that show similar expression patterns across a number of samples. The suggested metric for retrieval using clusterings is the normalized information distance. Empirical results finally suggest that inference for the full probabilistic model can be approximated with good performance using computationally faster heuristic clustering approaches (e.g. $k$-means). The method is highly scalable and straightforward to apply to construct a general-purpose gene expression experiment retrieval method.\\
\textbf{Availability:}
The method can be implemented using standard clustering algorithms and normalized information distance, available in many statistical software packages.
\end{abstract}

\footnotetext[1]{Shares equal contributions with the first author.}  
\footnotetext[2]{Currently at Institute for Adaptive and Neural Computation, School of Informatics, University of Edinburgh, UK}

\section{Introduction}\label{sec:intro}

As the use of high-throughput molecular measurement technologies continues to spread, an ever increasing amount of data from biological experiments is being stored in publicly available repositories. It is then often of interest for researchers to retrieve experimental datasets with relevance to a given experiment, in order to increase the power of statistical analyses and to be able to make novel findings not obtainable from one experiment alone. The current standard practice relies on searching for relevant experiments by keyword annotations \citep[e.g.][]{Zhu_et_al_2008}. However, despite efforts to maintain compliance with standard formats of documenting experiments, e.g. the MIAME standard \citep{Brazma_2001}, information about experiments may often be missing, insufficient or suffer from variations in terminology \citep[e.g.][]{Baumgartner_et_al_2007,Schmidberger_et_al_2011}. In view of the challenges associated with keyword-based retrieval, the complementary task of querying a database of experiments using measurement data, instead of keywords, has recently received increased attention in the literature.

Most earlier content-driven methods used for retrieval of gene expression data represent each experiment in terms of a profile over genes, or alternatively, over known gene sets or gene modules predicted from other data sources, see  \citet{Hunter_et_al_2001,Fujibuchi_et_al_2007,Caldas_et_al_2009,Engreitz_et_al_2010,Georgii_et_al_2012} and references therein. A representative example is to compute differential expression profiles of case vs. control, use the correlation between activity profiles as the measure of relevance, and retrieve the experiments with the highest correlations \citep[e.g.][]{Engreitz_et_al_2010}. This requires auxiliary information about the experiments, namely case and control labels of experiment samples, and possibly additional \emph{a priori} defined sets of important genes. In the context of gene expression time series, representative examples of retrieving gene expression profiles include \citet{Smith_et_al_2008} and \citet{Hafemeister_et_al_2011}.

Recently, two feasibility studies have gone beyond reducing experiments into single profiles by using probabilistic modelling of the experiments in the database being queried. \citet{Faisal_et_al_2014}, assumed that the query dataset can be explained as a mixture of the learnt models, each model learnt from one dataset, such that the measure of relevance is given by the inferred mixture weights. In a slightly different approach \citep{Seth_et_al_2014}, experiments were retrieved by evaluating the posterior marginal likelihoods, given the query data, of individual models stored for the experiments in the database. 

In this paper, we introduce a method for retrieving full datasets, i.e. experiments consisting of multiple samples, which is also based on probabilistic modelling. However, instead of using the query dataset itself as a query, we use a model learnt from it. The measure of relevance is therefore not a likelihood, but instead a suitably defined metric between the models. The argument is that for noisy and complex datasets, it is beneficial to extract relevant characteristics of the query dataset in the same way as was done with the datasets that are being queried. We also make explicit the importance of marginalizing out nuisance parameters which are not directly relevant for the retrieval task. For example, in a gene expression study, one is often more interested in how sets of genes are co-regulated, rather than their exact expression values which are additionally affected by numerous other influences. We tackle the specific problem of retrieving gene expression experiments by using a product partition model \citep{Jordan_et_al_2007} to cluster together genes that show similar expression patterns across a number of samples. By integrating out expression levels of the gene sets (i.e., cluster-specific information), only the co-expression patterns revealed by the clustering structure are retained. The clustering induced by the query dataset is then finally compared with the clusterings associated with the experiments in the database using the normalized information distance \citep{Vinh_et_al_2010}. Notice that this approach does not involve any ``training stage'', compared to that of \cite{Seth_et_al_2014}, and the retrieval step does not involve solving an optimization problem, compared to \cite{Faisal_et_al_2014}.

While gene clustering has a long history in characterizing gene expression datasets \citep{Eisen_et_al_1999,Dhaeseleer_2005}, it appears not to have been used in the context of experiment retrieval before. The use of gene clustering provides a straightforward way of characterizing each experiment with minimal preprocessing of the data while capturing central co-expression patterns. Essentially all previous approaches for retrieving gene expression data have converted the data to differential expression (or gene set enrichments) requiring fixed and known case-control distinctions. In contrast, we have only applied standard quality control and RMA normalization steps carried out in-house at the European Bioinformatics Institute (EBI) for datasets in the Expression Atlas database \citep[see][]{Petryszak_et_al_2014}. Our experimental evaluation further suggests that, for the current application, inference of the
full probabilistic model can be approximated by some computationally faster heuristic clustering algorithm, such as $k$-means (see Appendix \ref{app:simplified}). The computational simplicity makes the method highly scalable and easy to apply in a black-box manner, as a general-purpose retrieval scheme.

\section{Approach}\label{sec:approach}

Let $D_q$ denote a data matrix from some experiment of interest, and let $\{D_m\}_{m=1}^M$ be a database of $M$ datasets from previously conducted experiments. The aim is to retrieve datasets from among the $\{D_m\}_{m=1}^M$ with similar characteristics as the \emph{query} dataset $D_q$. Due to the complex nature of the data, there is no single sensible or obvious way of comparing datasets (matrices of possibly different sizes). We propose using a \emph{model} to characterize each dataset, with the aim of reducing noise and making relevant aspects of the data more tangible, while making the experiments comparable. The retrieval task then consists in ranking the models $\{\mathcal{M}_m\}_{m=1}^M$, inferred from $\{D_m\}_{m=1}^M$, with respect to their similarity with the query model $\mathcal{M}_q$ inferred from $D_q$. Note that in a broad sense, the commonly used differential expression can be considered as one model type, and clustering as another. 

To elaborate on the above idea further, we will now assume that the data generating mechanism of each dataset can be represented in terms of a probabilistic model with density $f$ in some family $\{f(\cdot|\theta)|\theta\in\Theta\}$. Often, the parameter $\theta$ can be decomposed as $\theta = (\lambda,\psi)$, where $\psi$ is the parameter or characteristic of interest (e.g., gene clusters) and $\lambda$ is a nuisance parameter (e.g., average expression level of the gene cluster). Marginalizing out (integrating the density over) $\lambda$ then yields a model family completely determined by $\psi\in\Psi$. 
Making this operation explicit, the key quantity used in inferring a representative model for a dataset $D$ is the \emph{marginal likelihood},
\begin{equation}\label{eq:marginal_likelihood}
p(D|\psi) = \int_{\Lambda}f(D|\lambda,\psi)\pi_{\lambda|\psi}(\lambda|\psi)\ud\lambda,
\end{equation}
where $\pi_{\lambda|\psi}(\cdot|\psi)$ is a prior density on $\Lambda$. Ideally, we would then proceed with a fully Bayesian approach to infer a posterior density (or distribution) $\pi_{\psi}(\cdot|D)$ over $\Psi$, and use $\mathcal{M}:=\pi_{\psi}(\cdot|D)$ to characterize $D$. 
However, for computational reasons we will here choose only a single element of $\Psi$ to represent $D$. Under zero-one loss, the optimal choice is then the \emph{maximum a posteriori} (MAP) solution 
\begin{equation}\label{eq:MAP}
\tilde{\psi} = \argmax_{\psi\in\Psi}\{p(D|\psi)\pi_{\psi}(\psi)\},
\end{equation} 
where $\pi_{\psi}$ is a prior over $\Psi$. 
Accordingly, we now define the representative model for $D$ as $\mathcal{M}:=\tilde{\psi}$.

If a suitable function $d:\mathpzc{M} \times \mathpzc{M}\rightarrow \R$ can be defined for the pairwise relations between the elements of the model space $\mathpzc{M}$, a natural ranking among $\mathcal{M}_1,\ldots,\mathcal{M}_M \in \mathpzc{M}$ will be induced by evaluating $d(\mathcal{M}_q,\mathcal{M}_m)$ for all $m$. For coherence of the ranking scheme, we will make a further assumption that $d$ is a \emph{metric}. That is, for all $\mathcal{M},\mathcal{M}',\mathcal{M}''\in \mathpzc{M}$, we require that  
\begin{equation*}
\begin{array}{rl}
(\text{M1}) & d(\mathcal{M},\mathcal{M}')\geq 0\\
(\text{M2}) & d(\mathcal{M},\mathcal{M}')= 0 \text{ if and only if } \mathcal{M}=\mathcal{M}'\\
(\text{M3}) & d(\mathcal{M},\mathcal{M}')=d(\mathcal{M}',\mathcal{M})\\
(\text{M4}) & d(\mathcal{M},\mathcal{M}'')\leq d(\mathcal{M},\mathcal{M}')+d(\mathcal{M}',\mathcal{M}'').
\end{array} 
\end{equation*}
With the above conditions satisfied, the function $d$ conforms to the intuition of a distance, and furthermore, provides a solid foundation for the design of data structures and algorithms, as the model space $\mathpzc{M}$ forms a metric space. We finally note that metrics are also available for probability distributions, making the described framework applicable in cases where computational resources allow for representing the elements of $\mathpzc{M}$ as full posterior distributions. 

\section{Methods}\label{sec:methods}

\subsection{Probabilistic model for gene clustering}
The first task in constructing a retrieval scheme is to choose an appropriate model for the experiments. While several different approaches, with varying aims and assumptions, exist for modelling gene expression data, a particularly simple and frequently used approach is that of gene clustering \citep[e.g.][]{Dhaeseleer_2005},  which seeks to cluster together genes that show similar expression patterns across a number of samples. Here, we use a probabilistic clustering approach which simultaneously infers both the number of clusters as well as the optimal clustering structure.   

Consider first a gene expression data matrix $D$ of dimension $n\times p$, where $n$ is the number of genes and $p$ is the number of samples. A \emph{clustering} $S=\{s_1\ldots,s_k\}$ is a partition of the set $N=\{1,\ldots,n\}$ into $k\in\{1,\ldots,n\}$ non-empty and non-overlapping subsets, or \emph{clusters}, such that $\cup_{c=1}^k s_c = N$ and $s_c\cap s_{c'}=\emptyset$, for $c\neq c'$. We focus here on a probabilistic formulation of clustering, which makes explicit use of partition structures, namely the \emph{product partition model} (PPM). Technically, PPM assumes that items in the same cluster are exchangeable and items in different clusters are independent \citep[see][]{Jordan_et_al_2007}. Using the terminology of Section \ref{sec:approach}, the parameter of interest for this model is the partition structure $S$, while the nuisance parameter is a vector of cluster-specific model parameters, $\boldsymbol{\lambda}=(\lambda_1,\ldots,\lambda_k)$. This leads to a marginal likelihood of the form (see Equation (\ref{eq:marginal_likelihood})) 
\begin{align}
&p(D|S)=\int_{\Lambda} f(D|\boldsymbol{\lambda},S)\pi_{\boldsymbol{\lambda}|S}(\boldsymbol{\lambda}|S)\ud\boldsymbol{\lambda}\nonumber\\
      =&\int_{\Lambda} \prod_{c=1}^k f\big(D^{(s_c)}|\lambda_c,s_c\big)\pi_{\lambda|S}(\lambda_c|s_c)\ud\boldsymbol{\lambda}
      =\prod_{c=1}^k p\big(D^{(s_c)}|s_c\big),\label{eq:PPM_likelihood}
\end{align}
where $D^{(s_c)}$ denotes the subset of $D$ which is indexed by $s_c$. Note that the assumption of independence between clusters entails constructing the marginal likelihood as a product of cluster-specific components. 

The prior distribution for $S$ will likewise be constructed as a product,
\begin{equation}\label{eq:PPM_prior}
\P(S) = K \prod_{c=1}^k h(s_c),\quad \text{for all }k\in\{1,\ldots,n\},
\end{equation}
where $K$ ensures normalization to 1 over the  model space $\mathcal{S}$ and $h(s_c)\geq 0$ for all subsets $s_c$. 
Note that (\ref{eq:PPM_prior}) actually specifies the joint distribution for $S$ and $k$, but since the latter is implied by the former, we omit $k$ from the notation. It can be shown that a PPM with $K$ and $h(s_c)$ chosen such that 
\begin{equation}\label{eq:CRP_prior}
\P(S) = \frac{\eta_0^k\prod_{c=1}^k(|s_c|-1)!}{\prod_{i=1}^n \eta_0+i-1},
\end{equation}
where $|s_c|$ is the number of observations in cluster $s_c$ and $\eta_0>0$ controls the tendency to form new clusters, can be obtained by  integrating out the model parameters in a Dirichlet process mixture model \citep{Dahl_2009}.

The cluster-specific marginal likelihoods $p\big(D^{(s_c)}|s_c\big)$ in Equation (\ref{eq:PPM_likelihood}) can in principle take any suitable form. Here, we assume that for $D^{(s_c)} = [x_{ij}]$, $i\in s_c$, $j = 1,\ldots,p$, the observations in each sample $j$ are independently generated from $\mathrm{N}\big(\mu_{cj},\tau_{cj}^{-1}\big)$ with a conjugate $\mathrm{NormalGamma}(\mu_0,\rho_0,\alpha_0,\beta_0)$ prior on the unknown model parameters. Furthermore, we make the simplistic assumption that the samples themselves are independent, conditional on a cluster assignment \citep[see][for a discussion about the implications of this assumption in a classification context]{Hand_Yu_2001}. The resulting cluster-specific marginal likelihoods may then be written as
\begin{equation}\label{eq:specific_marg_lklhd}
p\big(D^{(s_c)}|s_c\big)
= \prod_{j=1}^p (2\pi)^{-\frac{|s_c|}{2}}\Big(\frac{\rho_{0}}{\rho_{j}}\Big)^{\frac{1}{2}}
\frac{\Gamma(\alpha_{j})}{\Gamma(\alpha_{0})}
\frac{\beta_{0}^{\alpha_{0}}}{\beta_{j}^{\alpha_{j}}},
\end{equation}
where
\begin{align*}
\rho_{j}=&\;\rho_{0}+|s_c|,\;
\alpha_{j}=\,\alpha_{0}+\frac{|s_c|}{2},\;
\bar{x}_{j}=\;\frac{1}{|s_c|}\sum_{i\in s_c}(x_{ij})\\
\beta_{j}=&\;\beta_{0}+\frac{1}{2}\sum_{i\in s_c}(x_{ij}-\bar{x}_{j})^2
+\frac{|s_c|\rho_{0}(\bar{x}_{j}-\mu_{0})^2}{2\rho_{j}} \; .
\end{align*}

\citet{Blomstedt_et_al_2015} introduced a PPM for clustering mixed discrete and continuous data, where the continuous component was of form (\ref{eq:specific_marg_lklhd}). Following their implementation, we normalize each column of the data matrix $D = \cup_{c=1}^k D^{(s_c)}$ to have zero mean and unit variance, and set the hyperparameter values to $\mu_0=0$ and
$\rho_0=\alpha_0=\beta_0=1$. Furthermore, the model is equipped with a prior of the form (\ref{eq:CRP_prior}), with $\eta_0=1$.
Finally, combining Equations (\ref{eq:PPM_likelihood})--(\ref{eq:specific_marg_lklhd}), an optimal clustering $\tilde{S}$ w.r.t. a dataset $D$ is given by the MAP solution (see Equation (\ref{eq:MAP}))
\begin{equation}\label{eq:PPM_MAP}
\tilde{S} = \argmax_{S\in\mathcal{S}}\{p(D|S)\P(S)\}.
\end{equation}

\subsubsection{Inference}\label{sec:inference}

To find the optimal clustering $\tilde{S}\in \mathcal{S}$ as defined in Equation (\ref{eq:PPM_MAP}), we use a stochastic greedy search algorithm,  which moves in the model space by successive application of move, split and merge operators; for further details, see \citet{Blomstedt_et_al_2015}. 
While being more efficient for the optimization task than standard Markov chain Monte Carlo methods, for large amounts of data the algorithm still requires a considerable amount of computation time. To that end, some computational simplifications based on heuristic clustering procedures will be discussed in Appendix \ref{app:simplified}.

\subsection{Distance metric for clusterings}

Assuming now that each of the experiments in a database has been represented with a clustering $S\in\mathcal{S}$, the remaining task is to find a function $d$ which can be defined on $\mathcal{S}$ and satisfies conditions (M1)--(M4) above. In recent years, a new generation of information-theoretic distance measures has emerged \citep[see e.g.][]{Meila_2007,Vinh_et_al_2010}, which possess many desirable properties, such as the metric property, and which have been employed because of their strong mathematical foundation and ability to detect non-linear similarities.

\citet{Vinh_et_al_2010} conducted a systematic comparison of information-theoretic distance measures, concluding that the preferred ``general-purpose'' measure for comparing clusterings is the \emph{normalized information distance}, later denoted $d_{NID}$. To give a definition of this measure, we first introduce some notation. Briefly, for two clusterings $S$ and $S'$, the number of items co-occurring in clusters $s_c\in S$ and $s_{c'}\in S'$ is given by $n_{cc'} = |s_c\cap s'_{c'}|$, with $\sum_{c=1}^{k} \sum_{c'=1}^{k'} n_{cc'}=n$. The marginal sums are denoted by $n_{c\cdot}=\sum_{c'=1}^{k'} n_{cc'}$ and $n_{\cdot c'}= \sum_{c=1}^{k} n_{cc'}$. A key realization in the derivation of information-theoretic distance measures is that each clustering induces an empirical probability distribution over the set $\{1,\ldots,k\}$, such that the probability of a randomly chosen item $i\in N$ being in cluster $s_c$ is given by $\P(i\in s_c) = n_{c\cdot}/n$. Similarly, the joint probability of the pair $(i,j)\in N\times N$ co-occurring in clusters $s_c$ and $s'_{c'}$ is given by $\P\big((i,j)\in s_c \times s'_{c'}\big) = n_{cc'}/n$. The \emph{entropy} of a clustering $S$, describing the uncertainty associated with assigning items into the clusters of $S$, is then formulated as
\[
H(S) = -\sum_{c=1}^k \P(i\in s_c)\log \P(i\in s_c).
\]
The \emph{mutual information} of clusterings $S$ and $S'$, which measures how much having knowledge of $S'$ reduces $H(S)$ (or vice versa), is further defined as
\[
I(S,S') = \sum_{c=1}^k \sum_{c'=1}^{k'} \P\big((i,j)\in s_c \times s'_{c'}\big)
\log \frac{\P\big((i,j)\in s_c \times s'_{c'}\big)}{\P(i\in s_c) \P(j\in s'_{c'})}.
\]
It can also be interpreted as a measure of dependence in the sense that if $S$ and $S'$ are independent, then $I(S,S')=0$.  
Finally, from the above quantities we obtain $d_{NID}$ as
\begin{equation}\label{eq:nid}
d_{NID}(S,S') = 1-\frac{I(S,S')}{\max\{H(S),H(S')\}}.
\end{equation}

\section{Results}\label{sec:results}

\subsection{Data and experimental setup}
\label{sec:setup}

To evaluate the modelling-based retrieval scheme developed in Sections \ref{sec:approach} and \ref{sec:methods}, we used as a starting point all differential expression experiments conducted on the A-AFFY-44 affymetrix genechip available in Expression Atlas \citep[EA; \url{http://www.ebi.ac.uk/gxa}, see][]{Petryszak_et_al_2014} as of 4-Jun-2014. Only experiments with both measurement data and analytics data available were considered. Furthermore, experiments with a very small number of genes were discarded. Since most experiments had expression measurements for more than $54\,670$ genes, this number was set as the lower limit. Based on the above selection process we obtained an initial set of 447 experiments.
In a second stage, we selected a subset of these experiments based on the availability of experimental factor ontologies \citep[EFO; \url{http://www.ebi.ac.uk/efo/}, see][]{Malone03032010}, which were used as ground truth in the evaluation. 
More specifically, we retained those experiments which had at least one of the EFO types ``cell type'', ``disease'' or ``organism part'' present. Moreover, experiments having multiple values for a given EFO type were excluded, and finally only experiments with the same EFO value present in at least two experiments were included in this study, resulting in a final set of 251 experiments (for a list of accession numbers, see Appendix \ref{app:accession}). The number of samples per experiment varied between 6 and 353, the median number of samples being 22.

Out of the final set of 251 experiments, three partly ovelapping subsets corresponding to each of the EFO types were formed. These consisted of 103 experiments with values recorded for ``cell type'', 76 with values for ``disease'' and 174 with values for ``organism part''. The number of different EFO values in these sets of experiments were 23, 19 and 32, respectively. 
In retrieving full experiments, those experiments having the same EFO value were considered relevant, and other experiments irrelevant. Note that the above EFO types were not the main conditions of interest on which differential gene expression had been studied in the experiments, but were chosen to give a more general description of the experiments. A more complete ground truth was not readily available as most other EFO types were only present in small subsets of the experiments. Retrival performance was measured using precision and recall, taken as an average of successively using each of the experiments as a query to retrieve among the remaining experiments. 

In order to reduce the number of genes for clustering, we initially selected for each of the 251 experiments the top 5 genes resulting from a `non-specific' search in EA, in which genes with the highest absolute values of $t$-statistics in any available contrast come first, irrespective of whether they are reported with high $t$-statistics in the remaining contrasts \citep[for further details about listing genes in EA, see][]{Petryszak_et_al_2014}. Finally, by taking the union of these genes over all experiments, we arrived at $1125$ genes per experiment. The selection process \emph{per se} is not an essential part of our approach but done for computational convenience only. In a preliminary stage of our analyses, we experimented with different numbers of genes but found that this only had a minor impact on the results, see Appendix \ref{app:number_of_genes} for further details.

\subsection{Comparison of retrieval schemes}\label{sec:performance}

We will now proceed to evaluating the performance of the retrieval approach proposed in Section \ref{sec:approach}. For gene expression data, we learn for each experiment a Gaussian product partition model (PPM) which implies a clustering over genes, see Section \ref{sec:methods}. The clustering $S_q$ learned from the query data is then related to the clusterings $S_1,\ldots,S_M$ by evaluating the distances $d_{NID}(S_q,S_m)$, $m=1,\ldots,M$, see Equation (\ref{eq:nid}). 
This approach will be contrasted with two alternative approaches for content-based retrieval previously suggested in the literature. The first one of these is closely related to the proposed approach in that it learns a PPM for each experiment in the database. However, instead of evaluating  distances, it evaluates the marginal likelihoods $p(D_q|S_m)$ of the learnt models, given the query dataset. A higher likelihood is then an indication of a higher relevance to the query dataset. A similar approach, albeit for a different model family, was recently suggested in \citet{Seth_et_al_2014}. The term ``modelling-based retrieval'' has previously been used by \citet{Faisal_et_al_2014} to describe an approach based on probabilistic modelling but using a likelihood as the measure of relevance. To make a distinction between the approach proposed here and approaches based on evaluating likelihoods, we will in this comparison refer to the former as \emph{model-distance-based retrieval} and the latter as \emph{likelihood-based retrieval}. See Section \ref{sec:discussion} for a further discussion about the differences between the two approaches. 

The second alternative approach, \emph{differential expression based retrieval}, assumes that a statistical test to detect differentially expressed genes has been conducted beforehand. The method is then based on correlating the gene-specific differential expression $p$-values of the query experiment with those of the database experiments. An approach similar to this was suggested by \citet{Engreitz_et_al_2010}. If targeted at differential expression profiles obtained under specific conditions known to be important, this scheme has much potential to achieve good retrieval performance. On the other hand, it assumes more background knowledge and preprocessing of the data than the suggested retrieval schemes based on gene clustering. 
Here, we do not assume a specific condition of interest but choose in each experiment for the selected 1125 genes the smallest $p$-values under any of the conditions tested and reported in Expression Atlas. 
We also experimented with a much larger set of $40\,569$ genes, constituting the maximal common set of genes tested in all experiments, but this resulted in slightly inferior performance. The correlation measure used was Pearson's correlation. We finally note that differential expression based retrieval schemes can also be formulated under the general framework of Section \ref{sec:approach} using some appropriate probabilistic model for differential expression, as formulated in e.g. \citet{Do_et_al_2006}.

\begin{figure*}
        \centering
        \begin{subfigure}[b]{0.55\textwidth}
\includegraphics[width=\textwidth]{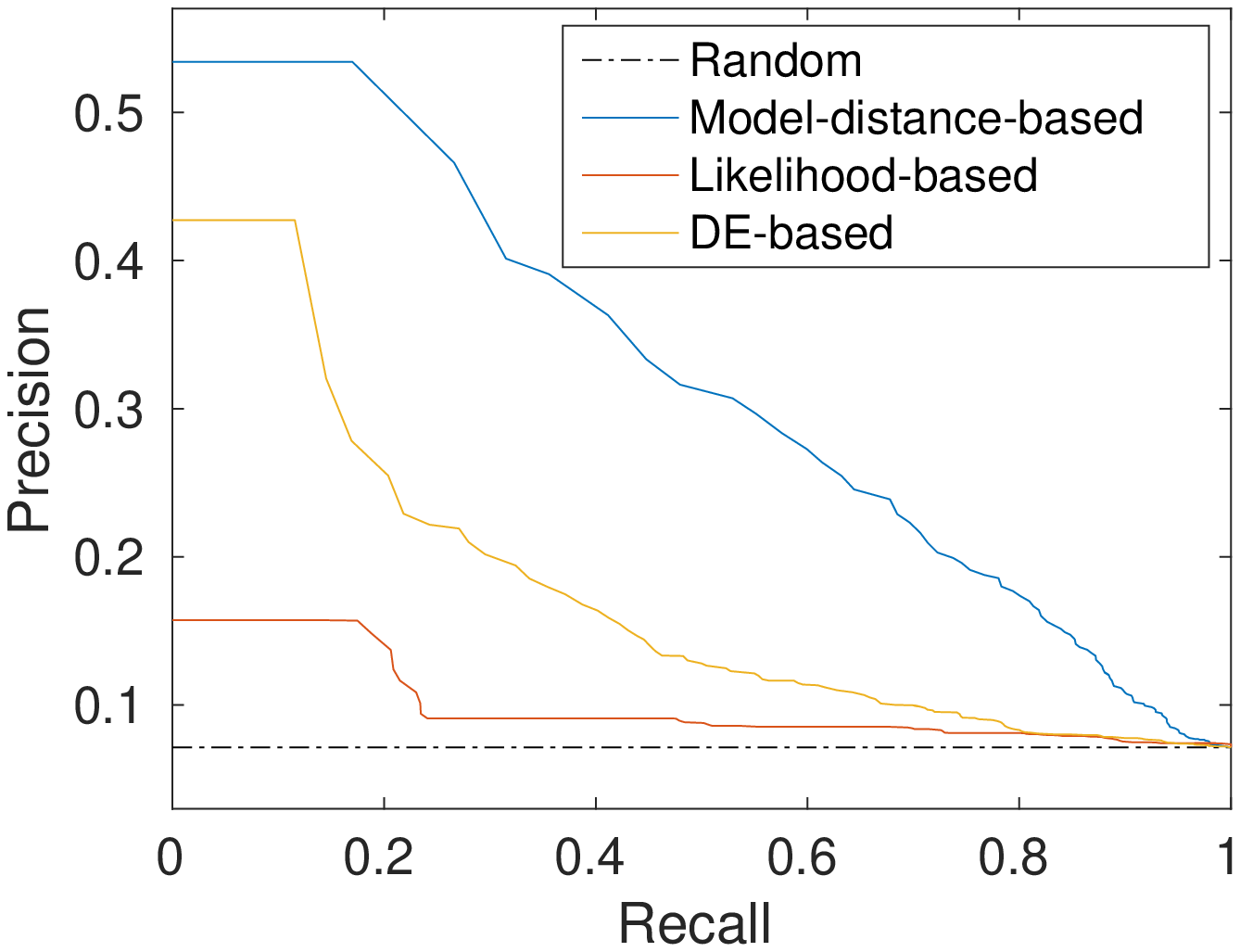}
\caption{Cell type}
        \end{subfigure}~
        \begin{subfigure}[b]{0.55\textwidth}
 \includegraphics[width=\textwidth]{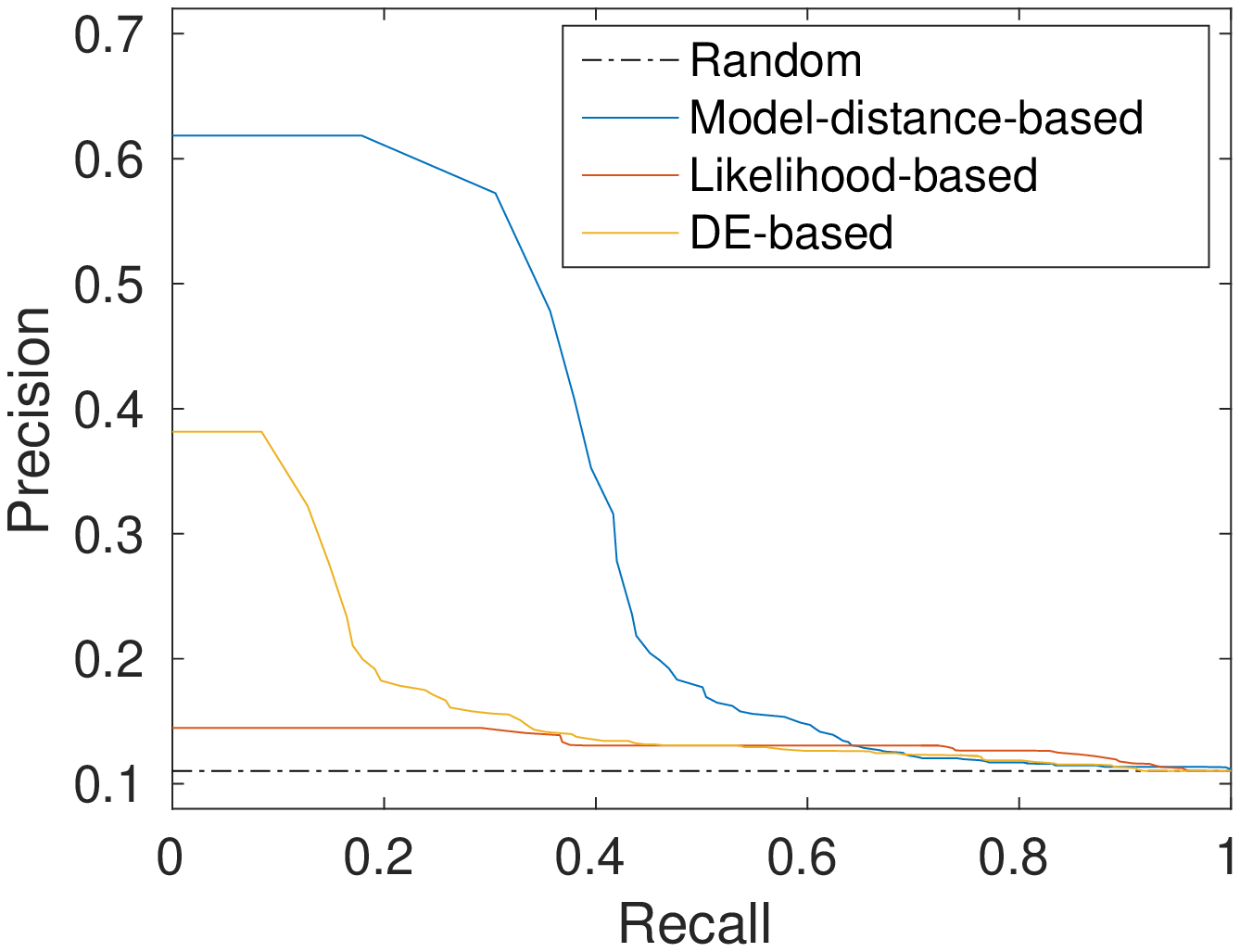}
 \caption{Disease}
        \end{subfigure}\\
        \begin{subfigure}[b]{0.55\textwidth}
\includegraphics[width=\textwidth]{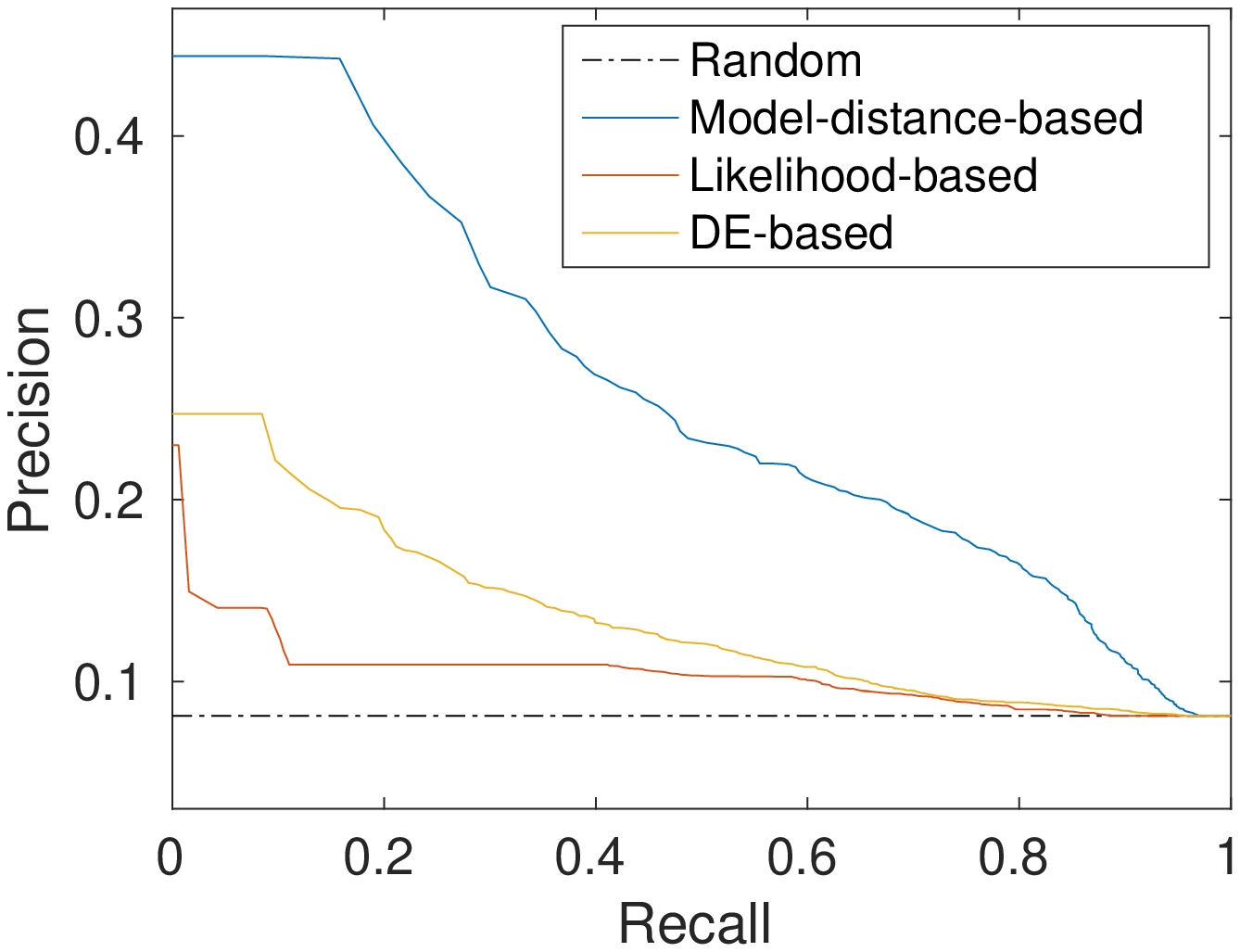}
\caption{Organism part}
        \end{subfigure}
        \caption{Precision-recall curves comparing model-distance-based, likelihood-based, and differential expression (DE) based retrieval using three EFO types (a--c) as ground truth.}
\label{fig:PR}
\end{figure*}

The results of the comparison between the retrieval schemes are shown in Figure \ref{fig:PR}. Here, the model-distance-based retrieval scheme clearly outperforms the two other schemes. A notable feature of the results is the surprisingly poor performance of the likelihood-based approach. This may be due to the well-known fact that gene expression measurements tend to be extremely noisy. In essence, the marginal likelihood $p(D_q|S_m)$ measures how well the query dataset $D_q$ is predicted by a model $S_m$, learnt from dataset $D_m$. Even if experiments $q$ and $m$ are in some way related, the idealized model $S_m$ may still not provide a good prediction for data $D_q$. Therefore, instead of using the complex and possibly very noisy dataset $D_q$ as query input, retaining only the characteristics relevant for retrieval in both $D_q$ and $D_m$ may help to improve performance, as illustrated in the results.

\subsection{Biological information in gene clustering}\label{sec:informativeness}

Any single EFO type will necessarily capture only one aspect of an experiment, whereas a meaningful retrieval task usually involves an evaluation of relevance between experiments in terms of a combination of aspects. It is therefore of interest to study the effect of composing the ground truth as a combination of multiple EFO types. In the current experimental setup, the ground truth for each of the EFO types ``cell type'', ``disease'' and ``organism part'', can be represented as a symmetric binary matrix $G$ of dimension $M\times M$, such that entry $g_{i,j}=1$ $\mathrm{iff}$ experiments $i$ and $j$ are mutually relevant. A ground truth which requires a match in $t$ EFO types can then be formed by summing the three matrices and requiring $g_{i,j}=t$.

\begin{figure*}
	\centering
	 \begin{subfigure}[b]{0.55\textwidth}
 \includegraphics[width=\textwidth]{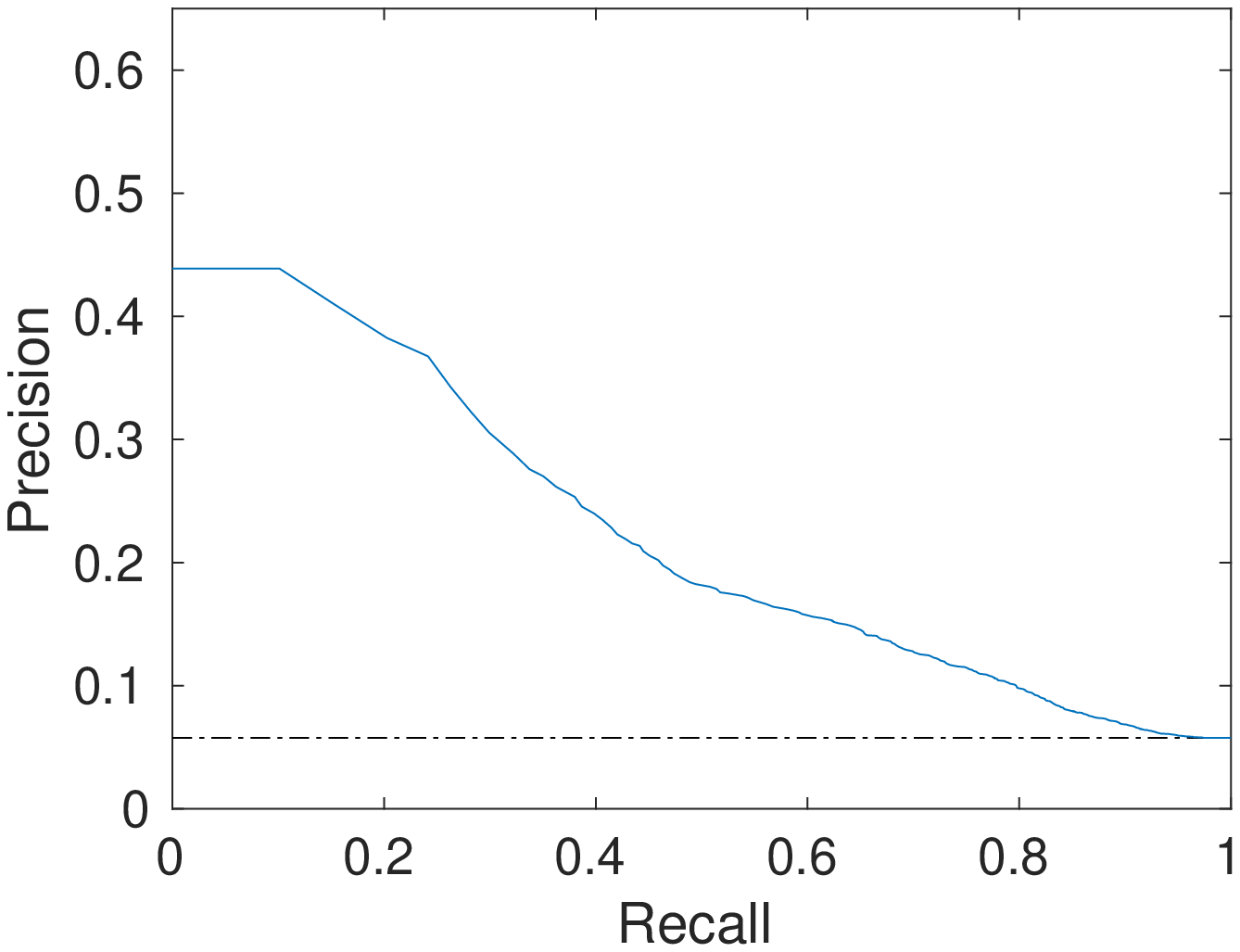}
 \caption{}
 \label{fig:simultaneous_a}
	\end{subfigure}~
	\begin{subfigure}[b]{0.55\textwidth}
\includegraphics[width=\textwidth]{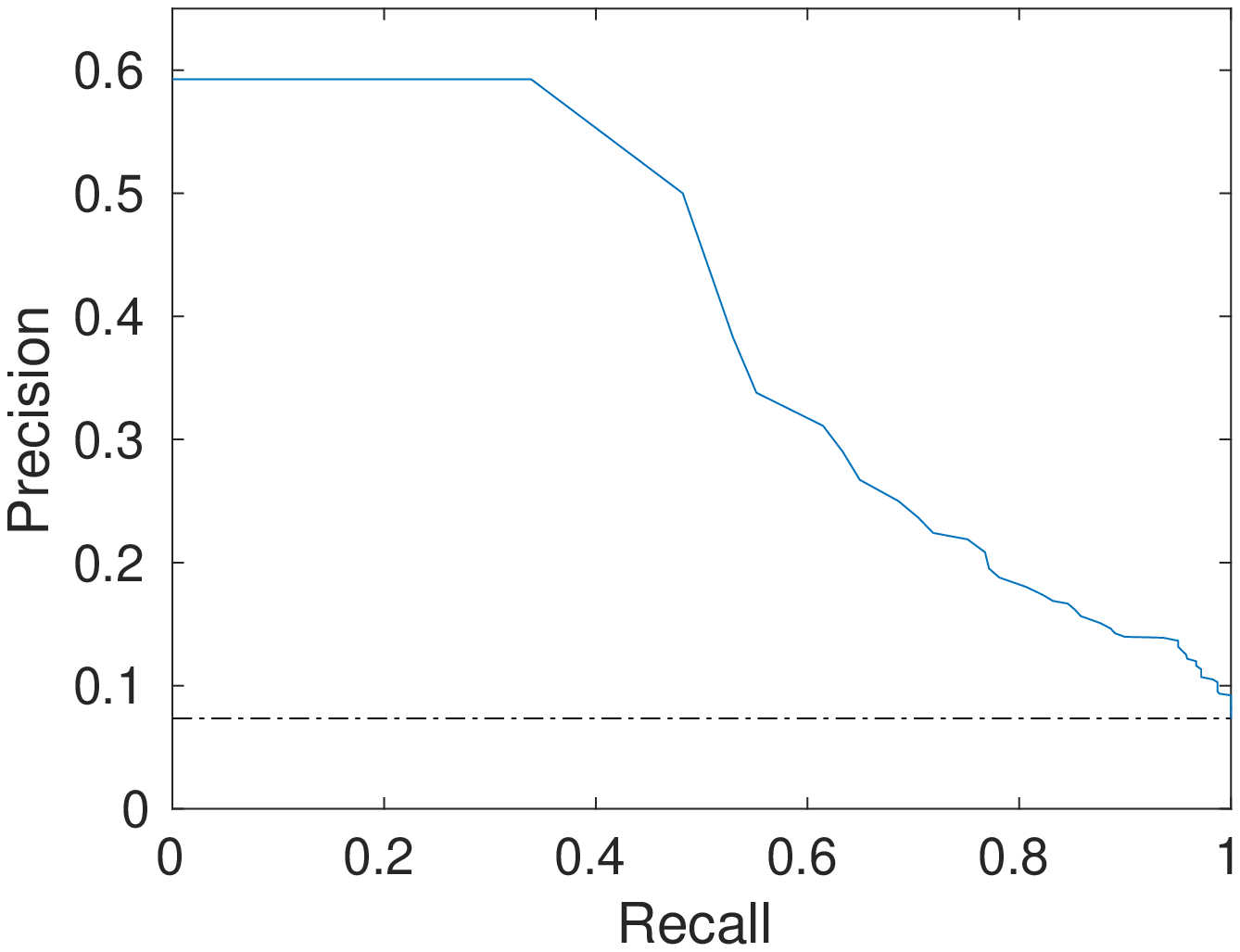}
\caption{}
\label{fig:simultaneous_b}
	\end{subfigure}\\
	\begin{subfigure}[b]{0.55\textwidth}
\includegraphics[width=\textwidth]{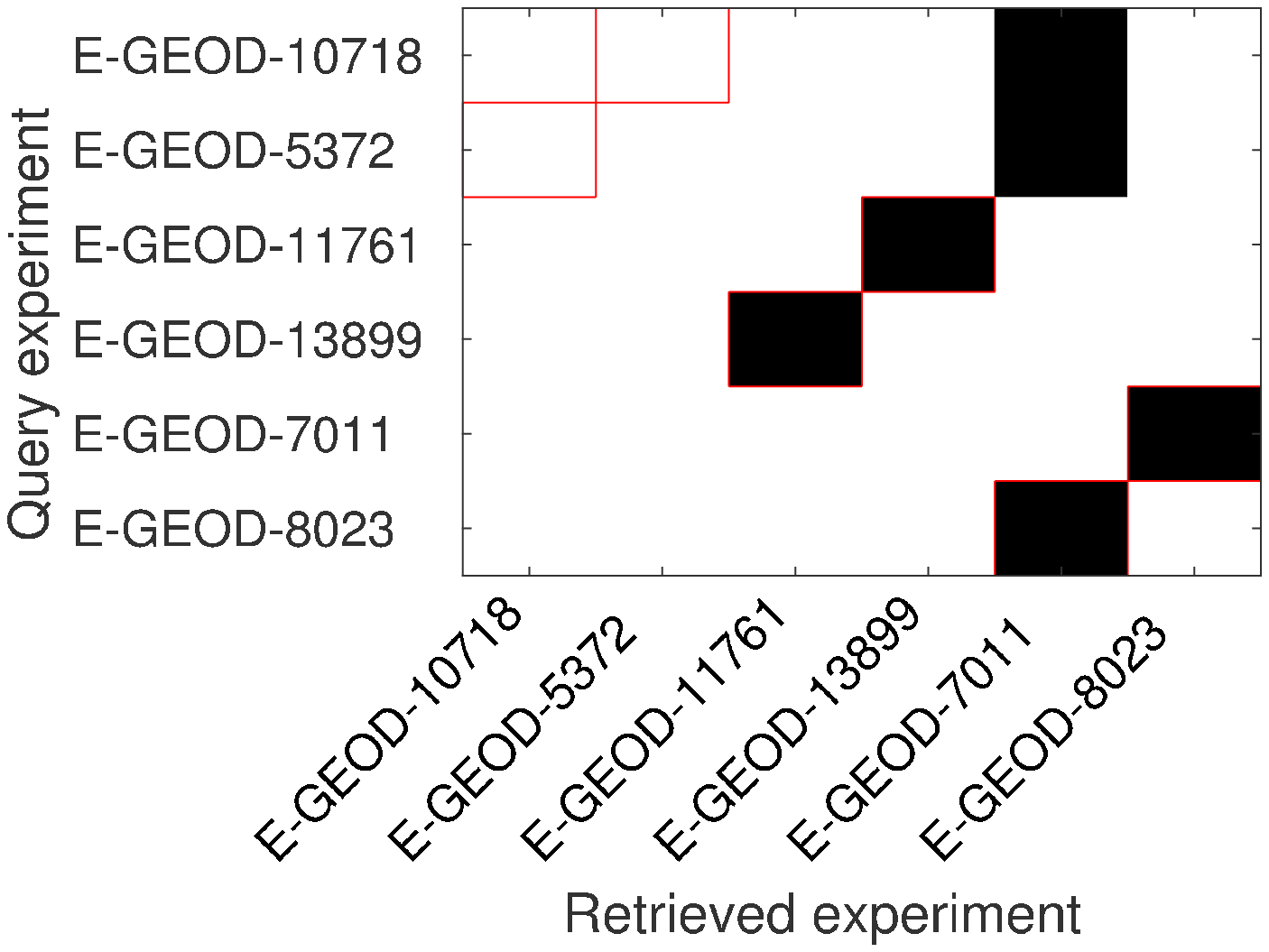}
\caption{}
\label{fig:simultaneous_c}
	\end{subfigure}
	\caption{Evaluation of model-distance-based retrieval scheme with respect to a ground truth requiring (a) at least
         one, (b) at least two, (c) exactly three matching EFO types. The rightmost subfigure compares the ground truth
         matrix (hollow squares) with the single most relevant retrieved experiment per query (solid squares) for the six
         experiments having a simultaneous match in all three EFO types. Accession numbers for the experiments are provided as a reference.}
\label{fig:simultaneous}
\end{figure*}

In Figure \ref{fig:simultaneous}, the model-distance-based retrieval scheme is evaluated against ground truth relevances requiring (a) any EFO type to match ($t\geq 1$) (b) two or more matches ($t \geq 2$) and (c) all EFO types to match ($t=3$). The number of experiments satisfying these conditions are 251, 54 and 6, respectively.  Intuitively, the ground truth can be considered increasingly informative as the number of matching EFO types required to declare relevance increases. A retrieval scheme capturing biologically relevant information should then be in better agreement with a more informative ground truth. Although the curves of Figures \ref{fig:simultaneous_a} and \ref{fig:simultaneous_b} are not directly comparable due to the differing number of experiments used, the shape of the latter gives an indication of a better agreement. In Figure \ref{fig:simultaneous_c}, owing to the small number of available experiments, the ground truth is compared with the single most relevant experiment (out of five possible ones) retrieved for each query. Here, the retrieval result matches the ground truth in four of the six queries.

\subsection{Annotations and gene clustering combined}\label{sec:annot_clust_combined}

As noted previously in Section \ref{sec:intro}, information about experiments may often be missing, insufficient or suffer from variations in terminology \citep[][]{Baumgartner_et_al_2007,Schmidberger_et_al_2011} despite a formal declaration of compliance with MIAME criteria \citep{Brazma_2001}. Hence, even in cases where keyword-based retrieval is of primary interest, it may be advantageous to complement a query with information provided by gene clustering. A straightforward way of combining these two types of information is the following. Assume that a database of $M$ experiments is being queried and that $L\leq M$ experiments are found to match the keyword query. More formally, the result can be encoded as a binary vector of length $M$ with $L$ elements having value 1. A model-distance-based retrieval scheme, on the other hand will return a vector of length $M$ with each element representing the distance of the corresponding experiment-specific model to the query model. Element-wise multiplication of these vectors then effectively induces a ranking of the experiments retrieved in the keyword-based query. The underlying idea is that this ranking will reflect some information which is not present in the queried keyword(s) alone.

To test the combined method, we considered all experiments matching in both ``cell type'' and ``organism part'', resulting in a total of 43 experiments (all other combinations of two EFO types resulted in significantly less experiments). A match in both of these EFO types was used as ground truth. The idea was then to retrieve experiments assuming only one of the EFO types to be known, complementing keyword-based retrieval with rankings from model-distance-based retrieval. Retrieving experiments assuming only ``cell type'' to be known resulted in an average precision of 0.55 for keyword-based retrieval and 0.61 (mean average precision) for combined retrieval, the corresponding numbers being 0.81 and 0.84, respectively, when only ``organism part'' was assumed to be known. In both cases we were able to see a slight improvement in performance for the combined approach, suggesting that keyword-based retrieval may benefit from being complemented with auxiliary information, such as gene clustering.

\section{Discussion}\label{sec:discussion}

In this paper, we have introduced a general probabilistic framework for content-driven retrieval of experimental datasets.  Compared to earlier works which also employ probabilistic modelling \citep[e.g.][]{Caldas_et_al_2009,Caldas_et_al_2012, Faisal_et_al_2014,Seth_et_al_2014}, we do not use the likelihood of the query data as a measure of relevance, but instead learn a model of the query data and compare models. We argue that this reduces noise in the query input. With nuisance parameters further marginalized out, only characteristics relevant for the retrieval task are retained. A special instance of the general framework introduced in this paper has been previously used as a comparative method in a simulation study \citep{Seth_et_al_2014} with performance slightly inferior to a likelihood-based approach. The simulation setting in that earlier study was, however, very simplistic compared to datasets encountered in many real-life scenarios, such as that of Section \ref{sec:results}, where the model-distance-based approach was now seen to clearly outperform its likelihood-based counterpart.

Contrary to likelihood-based approaches, the model-distance-based approach requires all models under consideration to belong to the same family. Although this may seem somewhat restrictive, in particular for the potential future scenario in which individual researchers independently store models in a repository along with their datasets \citep[e.g.][]{Faisal_et_al_2014}, there are also scenarios where the assumption is feasible. Datasets which arise as a result of some specific type of experiment are often in practice modelled using a fairly standardized set of approaches. In particular, if the models are constructed automatically, or by a curator of a data repository, the assumption of the models belonging to the same family is feasible.

As a specific application of the general framework, in Sections \ref{sec:methods} and \ref{sec:results} we proposed a retrieval scheme for gene expression experiments based on gene clustering. It turned out that clustering is even a surprisingly good model for this purpose; with minimal preprocessing and prior knowledge about the experiments, it is able to yield reasonable retrieval performance (Section \ref{sec:performance}) and to capture biologically relevant characteristics about the experiments (Section \ref{sec:informativeness}). Finally, we showed that it is straightforward to combine model-distance-based (or any modelling-based) retrieval with retrieval using available keywords (Section \ref{sec:annot_clust_combined}).

%
%

\section*{Acknowledgement}
The authors would like to thank Ugis Sarkans for providing useful information about Expression Atlas.  
This work was financially supported by the Academy of Finland (Finnish Centre of Excellence in Computational Inference Research COIN, grant no 251170).

\newpage

\appendix

\section*{Appendix}

\section{Simplified search for an optimal clustering}\label{app:simplified}

Recall that a product partition model (PPM) is a probabilistic model which implies a clustering $S=\{s_1\ldots,s_k\}$ of $n$ data items into $k\leq n$ non-empty and non-overlapping subsets. Given a dataset $D$, an optimal clustering 
$\tilde{S}$ is given by the \emph{maximum a posteriori} solution 
\begin{equation*}
\tilde{S} = \argmax_{S\in\mathcal{S}}\{p(D|S)\P(S)\},
\end{equation*}
where $\mathcal{S}$ denotes the space of all possible clusterings of $D$. Since the cardinality $|\mathcal{S}|$ of the model space grows very quickly with $n$, an exhaustive evaluation of all posterior probabilities $\P(S|D)\propto p(D|S)\P(S)$, $S\in\mathcal{S}$, is not feasible in practice (for instance, with $n=50$, we already have $|\mathcal{S}| = 1.8572 \times 10^{47}$). Therefore a stochastic greedy search algorithm was implemented in the analyses of Section 4 to find the optimal clustering for each dataset. While being more efficient for the optimization task than standard Markov chain Monte Carlo methods, for large amounts of data the algorithm still requires a considerable amount of computation time. 

One possible simplification is to restrict the search to a subset $\mathcal{S}^*\subset\mathcal{S}$ of the model space by choosing a set of potentially good solutions in advance, and then selecting the optimal solution among them as
\begin{equation} \label{eq:restricted_MAP}
\tilde{S}^* = \argmax_{S\in\mathcal{S}^*}\{p(D|S)\P(S)\}.
\end{equation}
A straightforward way of finding  a suitable $\mathcal{S}^*$ 
is to only consider solutions found by one or several different heuristic clustering algorithms. These algorithms are usually fast to execute but provide no measure of uncertainty regarding the obtained solution and require the number of clusters $k$ to be fixed in advance. Running such an algorithm for all values of $k\in\{1,\ldots,n\}$ will reduce the cardinality of the search space to  $|\mathcal{S}^*|=n$, which in many cases is small enough to enable an exhaustive evaluation of the posterior probabilities of all clusterings in $\mathcal{S}^*$. Even a combination of, say, $L$ different algorithms still yields a model space with a cardinality of only $|\mathcal{S}^*|=L\cdot n$. 

\begin{figure}
\centering
        \begin{subfigure}[b]{0.55\textwidth}
\includegraphics[width=\textwidth]{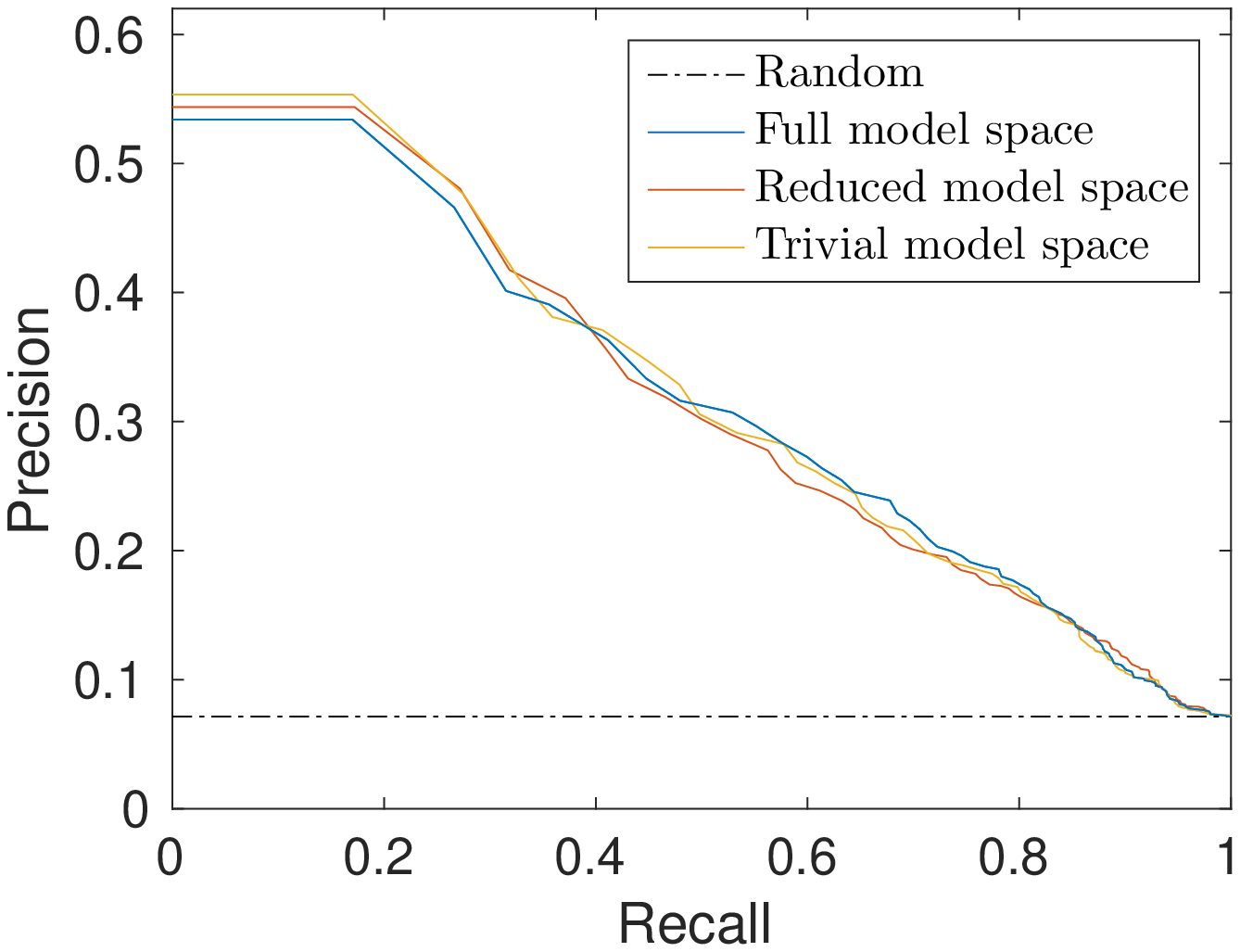}
\caption{Cell type}
        \end{subfigure}~
        \begin{subfigure}[b]{0.55\textwidth}
 \includegraphics[width=\textwidth]{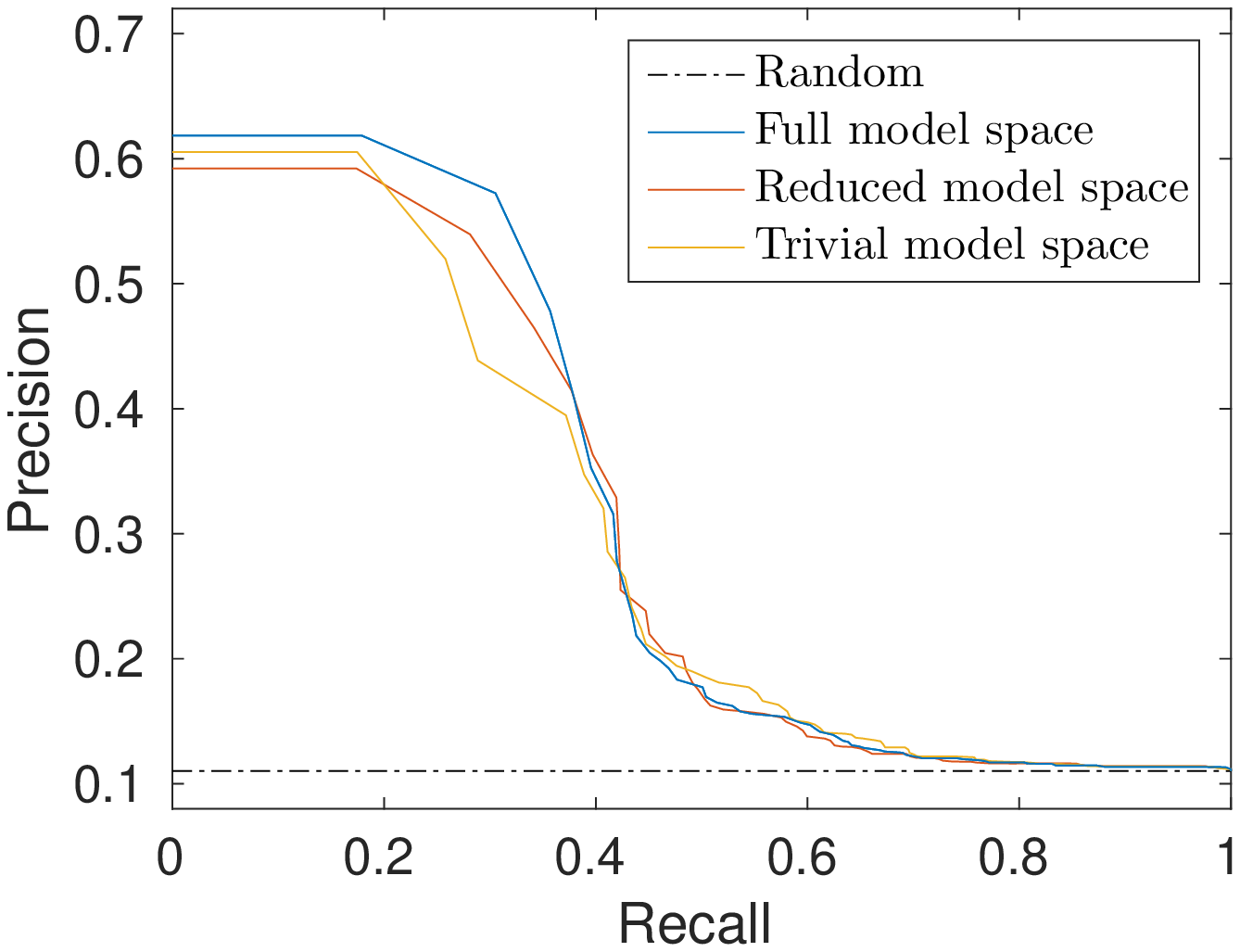}
 \caption{Disease}
        \end{subfigure}\\
        \begin{subfigure}[b]{0.55\textwidth}
\includegraphics[width=\textwidth]{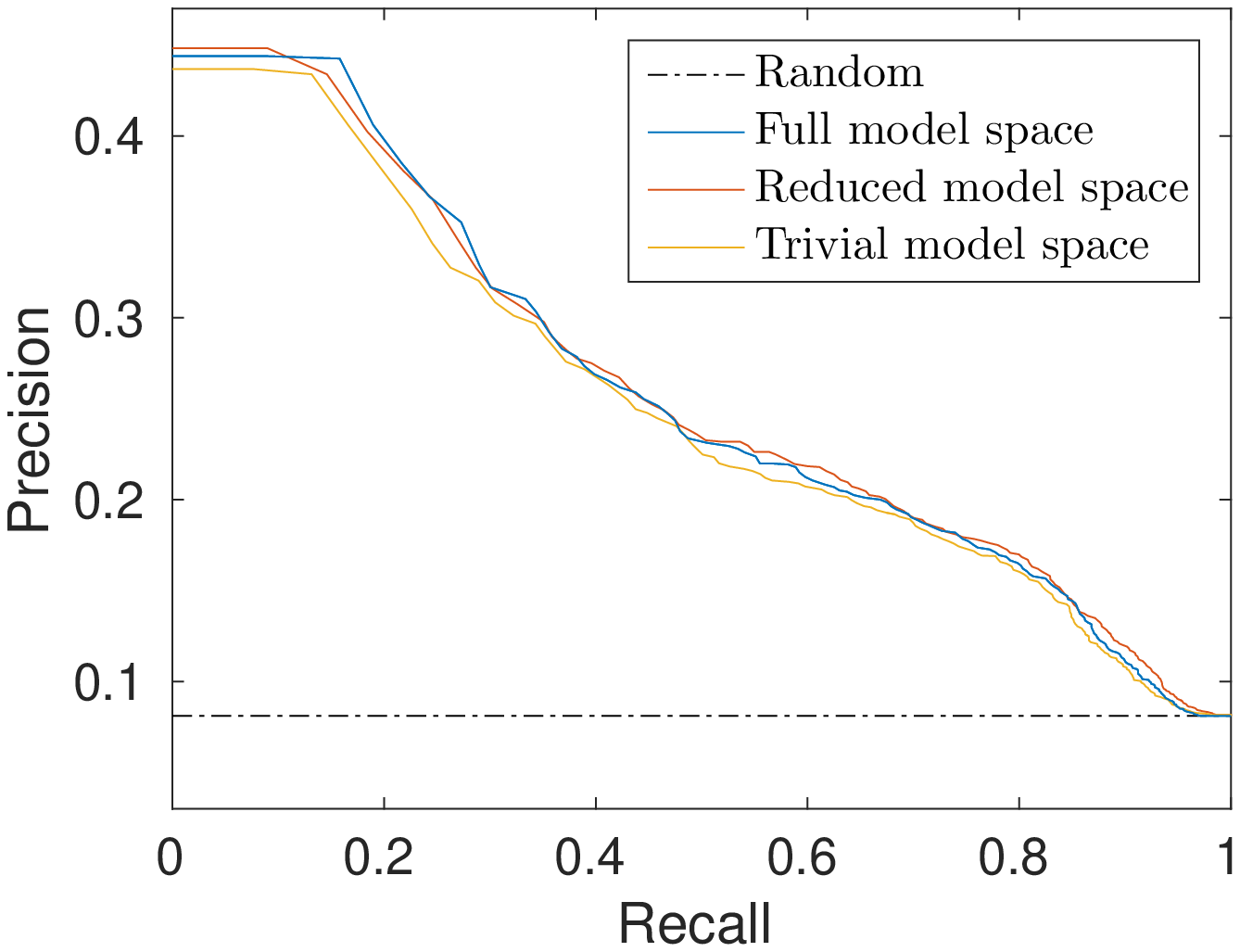}
\caption{Organism part}
        \end{subfigure}
        \caption{Retrieval performance using clusterings found in full $(\mathcal{S})$, reduced $(\mathcal{S}^*)$, and trivial $(\mathcal{S}^*_0)$ model spaces.}
\label{fig:modspace}
\end{figure}

To further reduce the scope of the search, the range of $k$ for which heuristic solutions are obtained may be restricted to an interval in which plausible solutions are likely to be found. For instance, in analysing how different distances and clustering methods interact regarding their ability to cluster gene expression, \citet{Jaskowiak_et_al_2014} conducted a comparison for clusterings generated in the interval $k\in\big\{2,\ldots,\big\lceil\sqrt{n}\big\rceil\big\}$, rather than the full range of values for $k$. In our current application, we additionally experimented with 
restricting $k$ to fixed value, which trivially reduces the model space to a single clustering. In this case, as the number of clusters is not chosen adaptively for each dataset, the clusterings no longer provide biologically meaningful groupings of the genes but may still give a sufficient characterization of the experiments for purposes of retrieval. This is demonstrated in Figure \ref{fig:modspace}, where retrieval based on the optimal clustering in the \emph{full model space} $\mathcal{S}$ is compared with that in a \emph{reduced model space} $\mathcal{S}^*$ of $k$-means solutions in $k\in\big\{2,\ldots,\big\lceil\sqrt{n}\big\rceil\big\}$, as well as a \emph{trivial model space} $\mathcal{S}^*_0$, consisting of only one $k$-means solution with $k=\lceil\sqrt{n}/2\rceil$, corresponding to the midpoint of the interval used by \citet{Jaskowiak_et_al_2014}. 

\begin{figure}
\centering
        \begin{subfigure}[b]{0.55\textwidth}
\includegraphics[width=\textwidth]{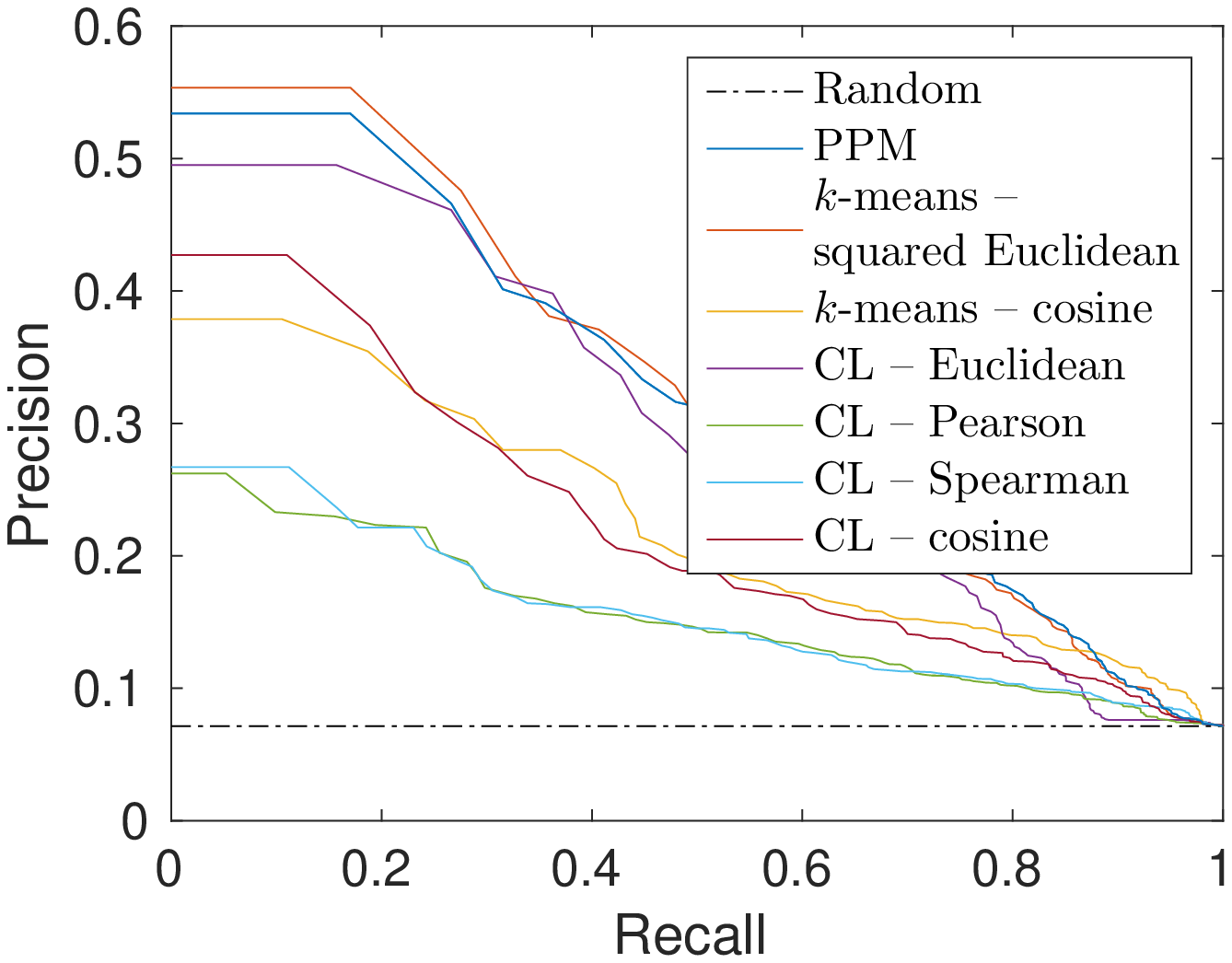}
\caption{Cell type}
        \end{subfigure}~
        \begin{subfigure}[b]{0.55\textwidth}
 \includegraphics[width=\textwidth]{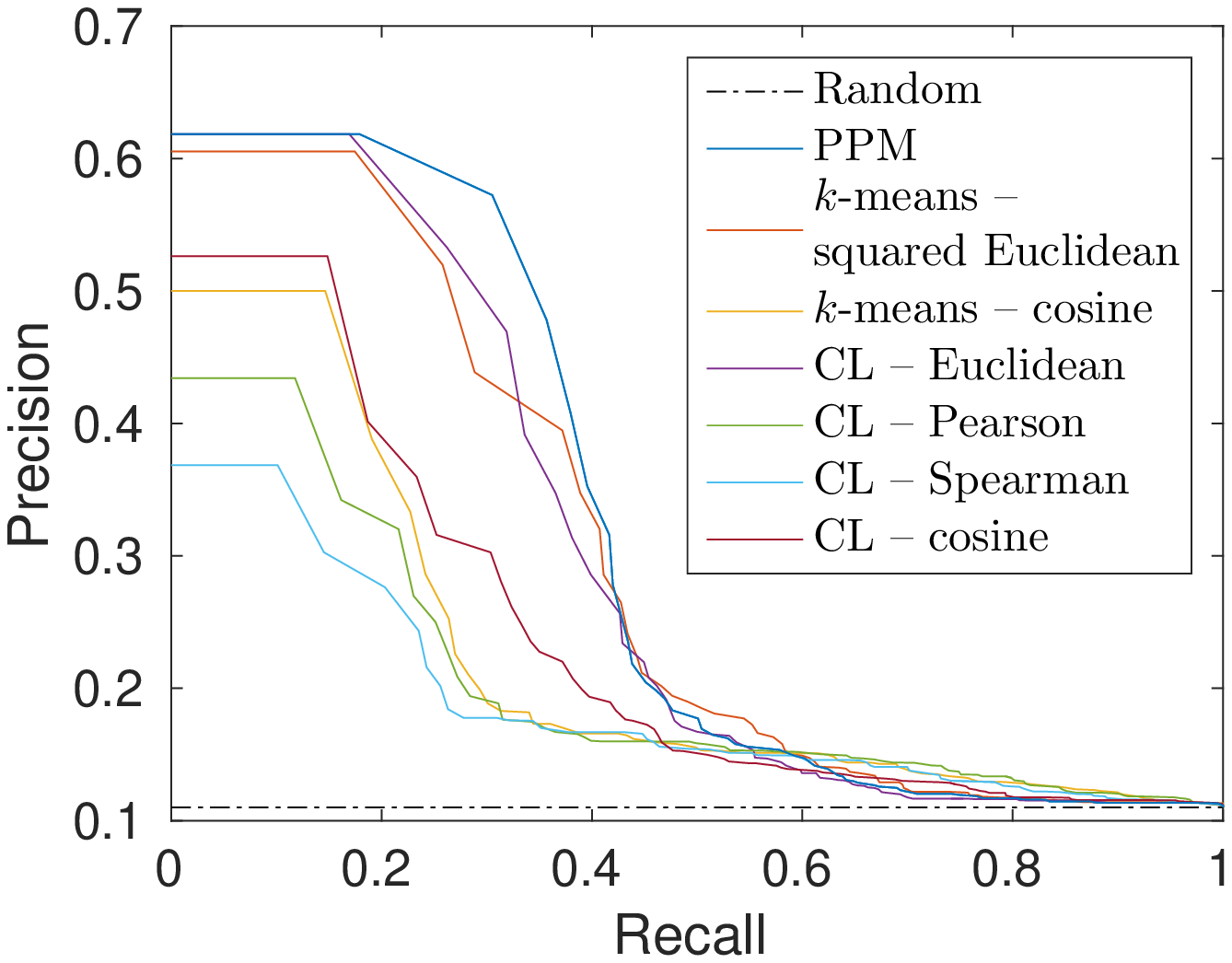}
 \caption{Disease}
        \end{subfigure}\\
        \begin{subfigure}[b]{0.55\textwidth}
\includegraphics[width=\textwidth]{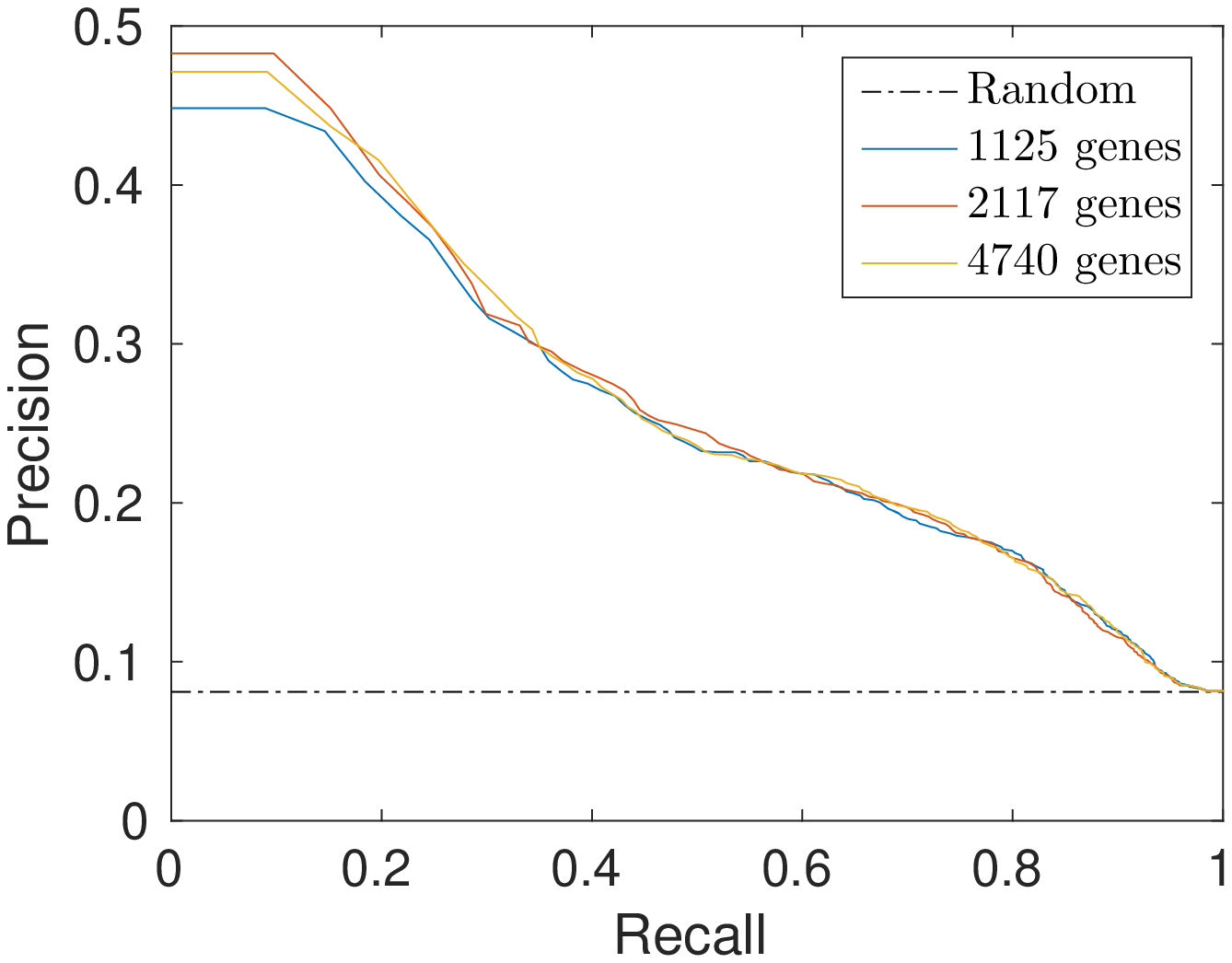}
\caption{Organism part}
        \end{subfigure}
        \caption{Retrieval performance for various heuristic clustering approaches, using PPM as baseline.}
\label{fig:algorithm}
\end{figure}

The quality of the solution in (\ref{eq:restricted_MAP}) depends on how well the clusterings in $\mathcal{S}^*$ (or $\mathcal{S}^*_0$) correspond to those clusterings in $\mathcal{S}$ which have a high probability under the PPM formulation. Figure \ref{fig:algorithm} shows a comparison of the retrieval performance of various heuristic clustering algorithms, with the number of clusters fixed for simplicity at $k=\lceil\sqrt{n}/2\rceil$, and using PPM as baseline. The results indicate that heuristic algorithms which are based on a Euclidean distance measure (e.g. $k$-means with squared Euclidean distance and complete linkage (CL) with Euclidean distance) yield retrieval performance which closest matches that obtained using the Gaussian PPM. Although similar behaviour may be expected in other datasets of the same type, the conclusion is, however, data-specific and should not be generalized beyond the scope of the current data without further validation.

\section{Impact of number of genes}\label{app:number_of_genes}

\begin{figure}
        \centering
        \begin{subfigure}[b]{0.55\textwidth}
\includegraphics[width=\textwidth]{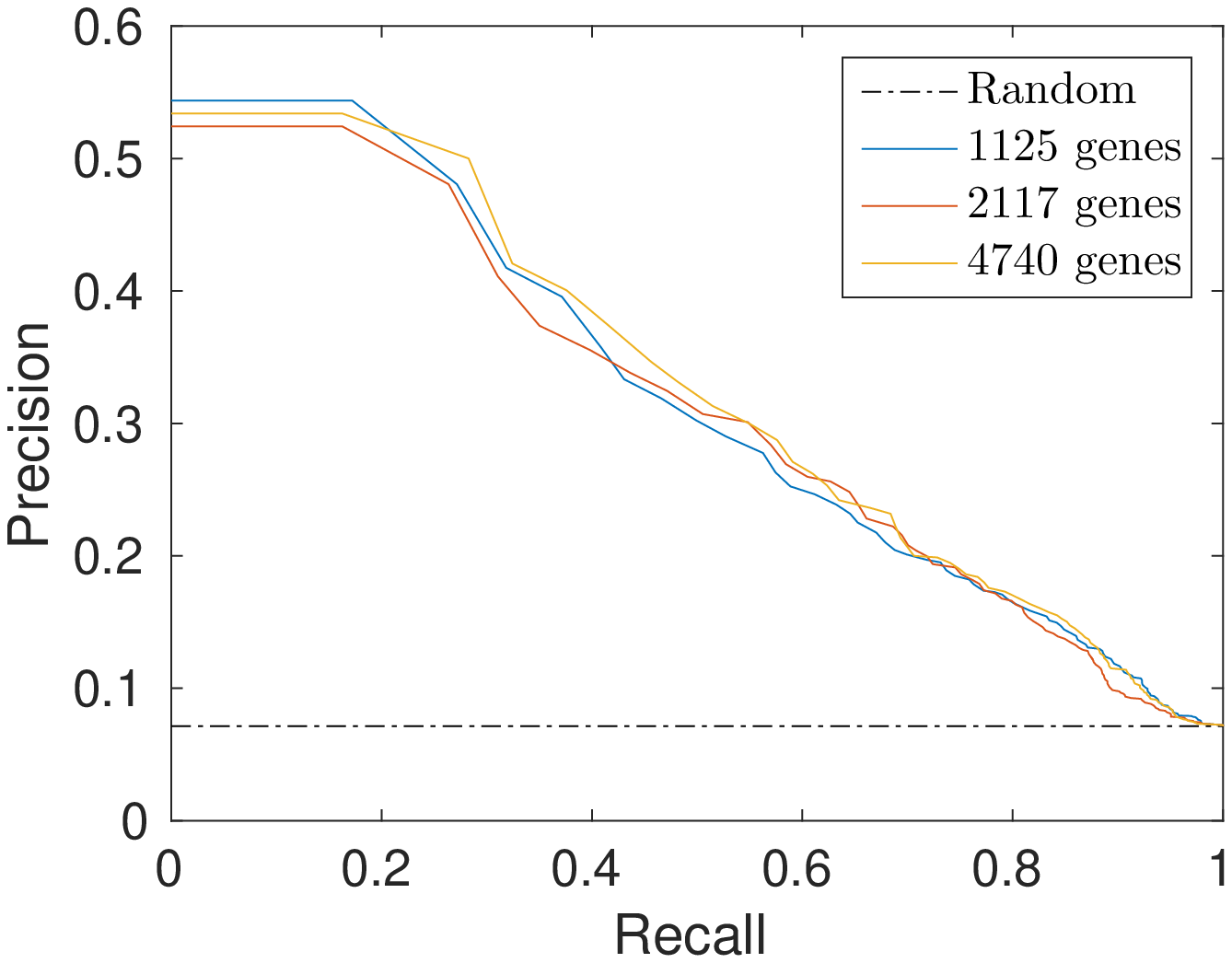}
\caption{Cell type}
        \end{subfigure}%
        \begin{subfigure}[b]{0.55\textwidth}
 \includegraphics[width=\textwidth]{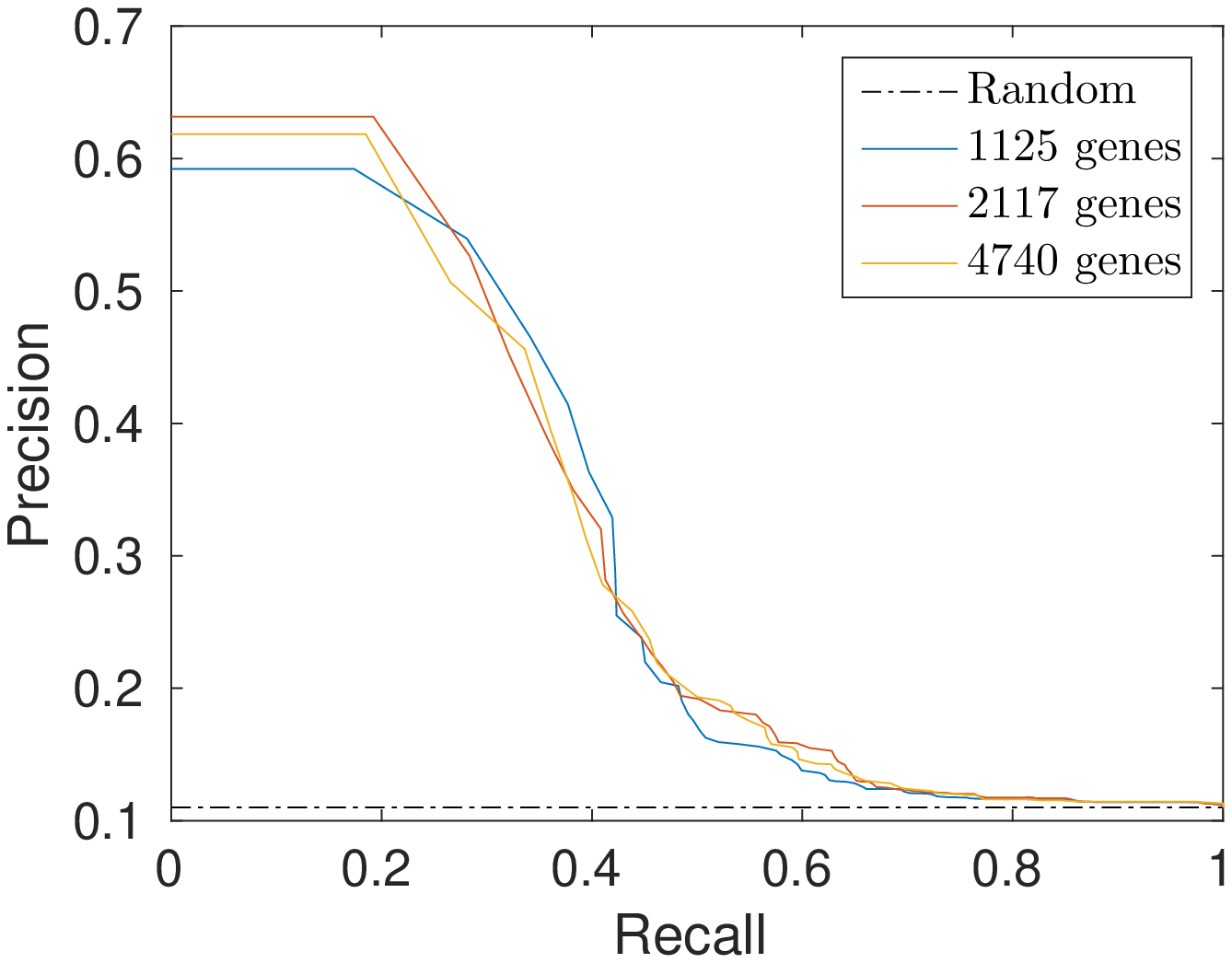}
 \caption{Disease}
        \end{subfigure}
        \begin{subfigure}[b]{0.55\textwidth}
\includegraphics[width=\textwidth]{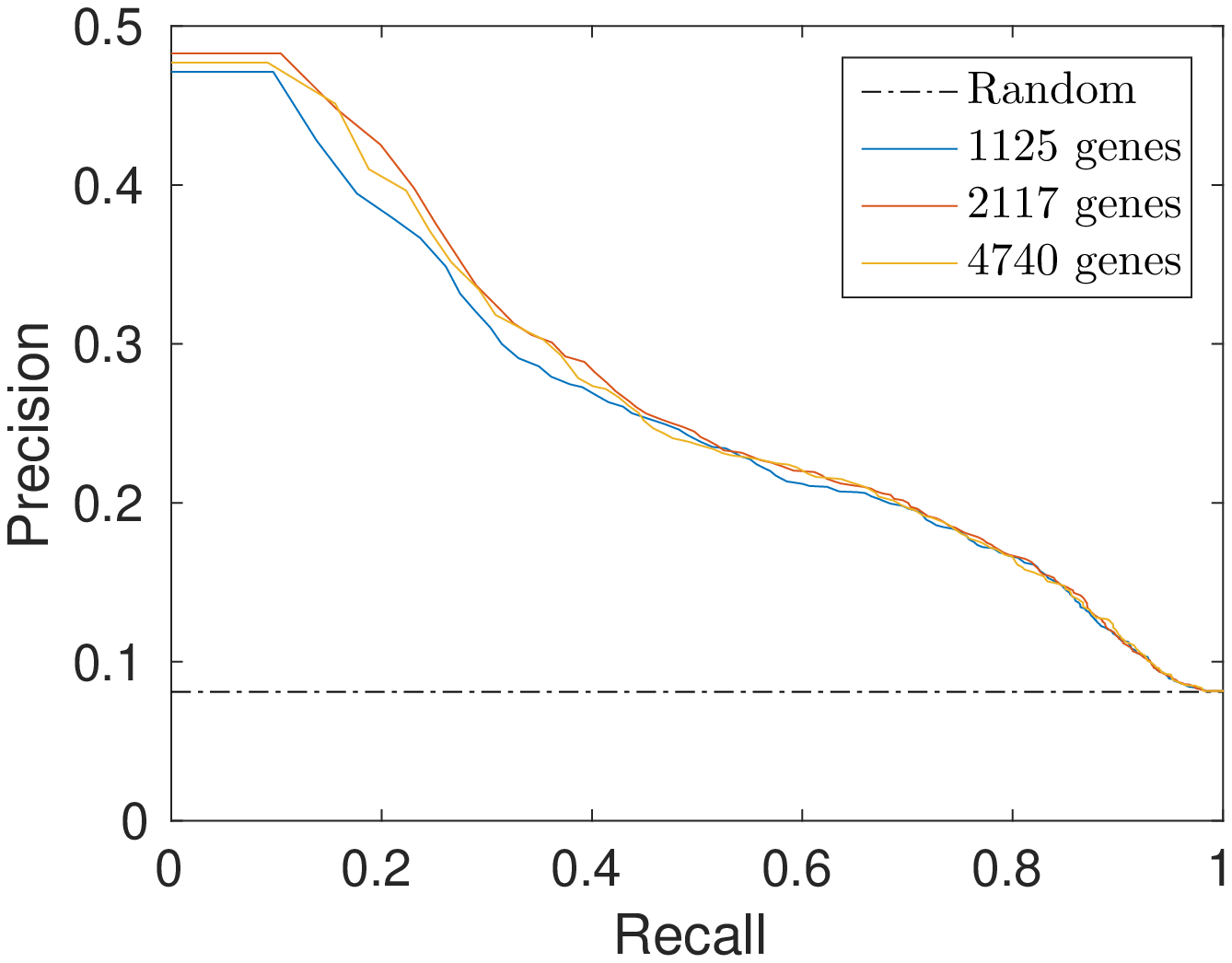}
\caption{Organism part}
        \end{subfigure}
        \caption{Retrieval performance for different numbers of genes included for clustering.}
\label{fig:ngenes}
\end{figure}

In Section 4.1, the number of genes for clustering was reduced by initially selecting for each experiment the top 5 genes resulting from a `non-specific' search in Expression Atlas \citep[\url{http://www.ebi.ac.uk/gxa}, see][]{Petryszak_et_al_2014}. Taking the union of these genes over all 251 experiments finally resulted in 1125 genes per experiment. To study the impact of the number of genes included in each dataset, we repeated the same procedure for the top 10 and top 25 genes, resulting in 2117 and 4740 genes per experiment, respectively. Due to the large number of genes, in particular in the last group, a simplified search scheme for clusterings was employed as described in the previous section, using $k$-means with squared Euclidean distance measure and 
$k\in\big\{2,\ldots,\big\lceil\sqrt{n}\big\rceil\big\}$. Figure \ref{fig:ngenes} suggests that the number of genes chosen only has a minor impact on retrieval performance.

\section{Experiment accession numbers}\label{app:accession}

Accession numbers of the 251 experiments selected for the analyses: 

\texttt{~\\
E-GEOD-10070, E-GEOD-10233, E-GEOD-10289, E-GEOD-10311, E-GEOD-10315, E-GEOD-10595, E-GEOD-10696, E-GEOD-10718, E-GEOD-10780, E-GEOD-10799, E-GEOD-10820, E-GEOD-10821, E-GEOD-10831, E-GEOD-10879, E-GEOD-10890, E-GEOD-10896, E-GEOD-10916, E-GEOD-10971, E-GEOD-10979, E-GEOD-11057, E-GEOD-11166, E-GEOD-11199, E-GEOD-11281, E-GEOD-11309, E-GEOD-11324, E-GEOD-11348, E-GEOD-11352, E-GEOD-11428, E-GEOD-11755, E-GEOD-11761, E-GEOD-11783, E-GEOD-11839, E-GEOD-11886, E-GEOD-11919, E-GEOD-11941, E-GEOD-11959, E-GEOD-12034, E-GEOD-12108, E-GEOD-12113, E-GEOD-12121, E-GEOD-12172, E-GEOD-12251, E-GEOD-12254, E-GEOD-12264, E-GEOD-12265, E-GEOD-12287, E-GEOD-12355, E-GEOD-12408, E-GEOD-12452, E-GEOD-12710, E-GEOD-13487, E-GEOD-13501, E-GEOD-13548, E-GEOD-13637, E-GEOD-13762, E-GEOD-13763, E-GEOD-13837, E-GEOD-13899, E-GEOD-13911, E-GEOD-13975, E-GEOD-13987, E-GEOD-14001, E-GEOD-14017, E-GEOD-14278, E-GEOD-14383, E-GEOD-14390, E-GEOD-14479, E-GEOD-14924, E-GEOD-14926, E-GEOD-14973, E-GEOD-15271, E-GEOD-15389, E-GEOD-15543, E-GEOD-15645, E-GEOD-15811, E-GEOD-15947, E-GEOD-16020, E-GEOD-16214, E-GEOD-16237, E-GEOD-16363, E-GEOD-1643, E-GEOD-16515, E-GEOD-16728, E-GEOD-16797, E-GEOD-16836, E-GEOD-16837, E-GEOD-17251, E-GEOD-17385, E-GEOD-17400, E-GEOD-17636, E-GEOD-17743, E-GEOD-17763, E-GEOD-18018, E-GEOD-18791, E-GEOD-18842, E-GEOD-18913, E-GEOD-18995, E-GEOD-19067, E-GEOD-19293, E-GEOD-19639, E-GEOD-19665, E-GEOD-19784, E-GEOD-19804, E-GEOD-19826, E-GEOD-19864, E-GEOD-19982, E-GEOD-20114, E-GEOD-20540, E-GEOD-20948, E-GEOD-21261, E-GEOD-22152, E-GEOD-22229, E-GEOD-22513, E-GEOD-22544, E-GEOD-22563, E-GEOD-22779, E-GEOD-23031, E-GEOD-23687, E-GEOD-23806, E-GEOD-2397, E-GEOD-23984, E-GEOD-25518, E-GEOD-2634, E-GEOD-26495, E-GEOD-26673, E-GEOD-27034, E-GEOD-2706, E-GEOD-27187, E-GEOD-31193, E-GEOD-32719, E-GEOD-3467, E-GEOD-34748, E-GEOD-34880, E-GEOD-3526, E-GEOD-35972, E-GEOD-36547, E-GEOD-3678, E-GEOD-3744, E-GEOD-3998, E-GEOD-4567, E-GEOD-4600, E-GEOD-4655, E-GEOD-4883, E-GEOD-4888, E-GEOD-5040, E-GEOD-5230, E-GEOD-5264, E-GEOD-5372, E-GEOD-5679, E-GEOD-6054, E-GEOD-6241, E-GEOD-6400, E-GEOD-6764, E-GEOD-7011, E-GEOD-7216, E-GEOD-7224, E-GEOD-7392, E-GEOD-7440, E-GEOD-7509, E-GEOD-7515, E-GEOD-7538, E-GEOD-7568, E-GEOD-7586, E-GEOD-7696, E-GEOD-7708, E-GEOD-7869, E-GEOD-7890, E-GEOD-8023, E-GEOD-8121, E-GEOD-8167, E-GEOD-8514, E-GEOD-8527, E-GEOD-8597, E-GEOD-8658, E-GEOD-8823, E-GEOD-8961, E-GEOD-8977, E-GEOD-9171, E-GEOD-9489, E-GEOD-9517, E-GEOD-9599, E-GEOD-9649, E-GEOD-9692, E-GEOD-9894, E-MEXP-1103, E-MEXP-1171, E-MEXP-1230, E-MEXP-1243, E-MEXP-1290, E-MEXP-1337, E-MEXP-1372, E-MEXP-1389, E-MEXP-1403, E-MEXP-1412, E-MEXP-1425, E-MEXP-1482, E-MEXP-1512, E-MEXP-1599, E-MEXP-1601, E-MEXP-1741, E-MEXP-1838, E-MEXP-1857, E-MEXP-1956, E-MEXP-1958, E-MEXP-2010, E-MEXP-2034, E-MEXP-2055, E-MEXP-2069, E-MEXP-2083, E-MEXP-2115, E-MEXP-2128, E-MEXP-2236, E-MEXP-2340, E-MEXP-2351, E-MEXP-2360, E-MEXP-2375, E-MEXP-2590, E-MEXP-2657, E-MEXP-3479, E-MEXP-3577, E-MEXP-3756, E-MEXP-3810, E-MEXP-555, E-MEXP-561, E-MEXP-563, E-MEXP-858, E-MEXP-884, E-MEXP-930, E-MEXP-935, E-MEXP-964, E-MEXP-980, E-MEXP-987, E-MEXP-993, E-MTAB-1131, E-MTAB-317, E-MTAB-372, E-MTAB-874, E-TABM-1020, E-TABM-1029, E-TABM-1138, E-TABM-1208, E-TABM-234, E-TABM-276, E-TABM-282, E-TABM-311, E-TABM-440, E-TABM-577, E-TABM-601, E-TABM-666, E-TABM-763, E-TABM-898
}


\begin{thebibliography}{}

\bibitem[Baumgartner {\em et~al.}(2007)Baumgartner, Cohen, Fox, Acquaah-Mensah,
  and Hunter]{Baumgartner_et_al_2007}
Baumgartner, {Jr.}, W.~A., Cohen, K.~B., Fox, L.~M., Acquaah-Mensah, G., and
  Hunter, L. (2007).
\newblock Manual curation is not sufficient for annotation of genomic
  databases.
\newblock {\em Bioinformatics\/}, {\bf 23}, i41--i48.

\bibitem[Blomstedt {\em et~al.}(2015)Blomstedt, Tang, Xiong, Granlund, and
  Corander]{Blomstedt_et_al_2015}
Blomstedt, P., Tang, J., Xiong, J., Granlund, C., and Corander, J. (2015).
\newblock A {B}ayesian predictive model for clustering data of mixed discrete
  and continuous type.
\newblock {\em IEEE Transactions on Pattern Analysis and Machine
  Intelligence\/}, {\bf 37}(3), 489--498.

\bibitem[Brazma(2001)Brazma]{Brazma_2001}
Brazma, A. (2001).
\newblock Minimum information about a microarray experiment ({MIAME}) --
  towards standards for microarray data.
\newblock {\em Nature Genetics\/}, {\bf 29}, 365--71.

\bibitem[Caldas {\em et~al.}(2009)Caldas, Gehlenborg, Faisal, Brazma, and
  Kaski]{Caldas_et_al_2009}
Caldas, J., Gehlenborg, N., Faisal, A., Brazma, A., and Kaski, S. (2009).
\newblock Probabilistic retrieval and visualization of biologically relevant
  microarray experiments.
\newblock {\em Bioinformatics\/}, {\bf 25}(12), i145--i153.

\bibitem[Caldas {\em et~al.}(2012)Caldas, Gehlenborg, Faisal, R{\"o}nty,
  Nicholson, Knuutila, Brazma, and Kaski]{Caldas_et_al_2012}
Caldas, J., Gehlenborg, Kettunen, E., Faisal, A., R{\"o}nty, M., Nicholson,
  A.~G., Knuutila, S., Brazma, A., and Kaski, S. (2012).
\newblock Data-driven information retrieval in heterogeneous collections of
  transcriptomics data links {SIM2s} to malignant pleural mesothelioma.
\newblock {\em Bioinformatics\/}, {\bf 28}(2), 246--253.

\bibitem[Dahl(2009)Dahl]{Dahl_2009}
Dahl, D.~B. (2009).
\newblock Modal clustering in a class of product partition models.
\newblock {\em Bayesian Analysis\/}, {\bf 4}(2), 243--264.

\bibitem[D'haeseleer(2005)D'haeseleer]{Dhaeseleer_2005}
D'haeseleer, P. (2005).
\newblock How does gene expression clustering work?
\newblock {\em Nature Biotechnology\/}, {\bf 23}(12), 1499--1501.

\bibitem[Do {\em et~al.}(2006)Do, M{\"u}ller, and Vannucci]{Do_et_al_2006}
Do, K.-A., M{\"u}ller, P., and Vannucci, M., editors (2006).
\newblock {\em Bayesian Inference for Gene Expression and Proteomics\/}.
\newblock Cambridge University Press, Cambridge, UK.

\bibitem[Eisen {\em et~al.}(1999)Eisen, Spellman, Brown, and
  Botstein]{Eisen_et_al_1999}
Eisen, M.~B., Spellman, P.~T., Brown, P.~O., and Botstein, D. (1999).
\newblock Cluster analysis and display of genome-wide expression patterns.
\newblock {\em PNAS\/}, {\bf 95}, 14863--14868.

\bibitem[Engreitz {\em et~al.}(2010)Engreitz, Morgan, Dudley, Chen, Thathoo,
  Altman, and Butte]{Engreitz_et_al_2010}
Engreitz, J.~M., Morgan, A.~A., Dudley, J.~T., Chen, R., Thathoo, R., Altman,
  R.~B., and Butte, A.~J. (2010).
\newblock Content-based microarray search using differential expression
  profiles.
\newblock {\em BMC Bioinformatics\/}, {\bf 11}(603).

\bibitem[Faisal {\em et~al.}(2014)Faisal, Peltonen, Georgii, Rung, and
  Kaski]{Faisal_et_al_2014}
Faisal, A., Peltonen, J., Georgii, E., Rung, J., and Kaski, S. (2014).
\newblock Toward computational cumulative biology by combining models of
  biological datasets.
\newblock {\em PLoS ONE\/}, {\bf 9}(11), e113053.

\bibitem[Fujibuchi {\em et~al.}(2007)Fujibuchi, Kiseleva, Taniguchi, Harada,
  and Horton]{Fujibuchi_et_al_2007}
Fujibuchi, W., Kiseleva, L., Taniguchi, T., Harada, H., and Horton, P. (2007).
\newblock Cellmontage: similar expression profile search server.
\newblock {\em Bioinformatics\/}, {\bf 23}(22), 3103--3104.

\bibitem[Georgii {\em et~al.}(2012)Georgii, Saloj{\"a}rvi, Brosch{\'e},
  Kangasj{\"a}rvi, and Kaski]{Georgii_et_al_2012}
Georgii, E., Saloj{\"a}rvi, J., Brosch{\'e}, M., Kangasj{\"a}rvi, J., and
  Kaski, S. (2012).
\newblock Targeted retrieval of gene expression measurements using regulatory
  models.
\newblock {\em Bioinformatics\/}, {\bf 28}(18), 2349--2356.

\bibitem[Hafemeister {\em et~al.}(2011)Hafemeister, Costa, Schonhuth, and
  Schliep]{Hafemeister_et_al_2011}
Hafemeister, C., Costa, I.~G., Schonhuth, A., and Schliep, A. (2011).
\newblock Classifying short gene expression time-courses with {B}ayesian
  estimation of piecewise constant functions.
\newblock {\em Bioinformatics\/}, {\bf 27}(7), 946--952.

\bibitem[Hand and Yu(2001)Hand and Yu]{Hand_Yu_2001}
Hand, D.~J. and Yu, K. (2001).
\newblock Idiot's {B}ayes -- {N}ot so stupid after all?
\newblock {\em International Statistical Review\/}, {\bf 69}(3), 385--398.

\bibitem[Hunter {\em et~al.}(2001)Hunter, Taylor, Leach, and
  Simon]{Hunter_et_al_2001}
Hunter, L., Taylor, R.~C., Leach, S.~M., and Simon, R. (2001).
\newblock {GEST}: a gene expression search tool based on a novel {B}ayesian
  similarity metric.
\newblock {\em Bioinformatics\/}, {\bf 17}(Suppl 1), S115--S122.

\bibitem[Jaskowiak {\em et~al.}(2014)Jaskowiak, Campello, and
  Costa]{Jaskowiak_et_al_2014}
Jaskowiak, P.~A., Campello, R. J. G.~B., and Costa, I.~G. (2014).
\newblock On the selection of appropriate distances for gene expression data
  clustering.
\newblock {\em BMC Bioinformatics\/}, {\bf 15}(Suppl 2), S2.

\bibitem[Jordan {\em et~al.}(2007)Jordan, Livingstone, and
  Barry]{Jordan_et_al_2007}
Jordan, C., Livingstone, V., and Barry, D. (2007).
\newblock Statistical modelling using product partition models.
\newblock {\em Statistical Modelling\/}, {\bf 7}(3), 275--295.

\bibitem[Malone {\em et~al.}(2010)Malone, Holloway, Adamusiak, Kapushesky,
  Zheng, Kolesnikov, Zhukova, Brazma, and Parkinson]{Malone03032010}
Malone, J., Holloway, E., Adamusiak, T., Kapushesky, M., Zheng, J., Kolesnikov,
  N., Zhukova, A., Brazma, A., and Parkinson, H. (2010).
\newblock Modeling sample variables with an experimental factor ontology.
\newblock {\em Bioinformatics\/}, {\bf 26}(8), 1112--1118.

\bibitem[Meil{\u a}(2007)Meil{\u a}]{Meila_2007}
Meil{\u a}, M. (2007).
\newblock Comparing clusterings--an information based distance.
\newblock {\em Journal of Multivariate Analysis\/}, {\bf 98}, 873--895.

\bibitem[Petryszak {\em et~al.}(2014)Petryszak, Burdett, Fiorelli, Fonseca,
  Gonzalez-Porta, Hastings, Huber, Jupp, Keays, Kryvych, McMurry, Marioni,
  Malone, Megy, Rustici, Tang, Taubert, Williams, Mannion, Parkinson, and
  Brazma]{Petryszak_et_al_2014}
Petryszak, R., Burdett, T., Fiorelli, B., Fonseca, N., Gonzalez-Porta, M.,
  Hastings, E., Huber, W., Jupp, S., Keays, M., Kryvych, N., McMurry, J.,
  Marioni, J., Malone, J., Megy, K., Rustici, G., Tang, A.~Y., Taubert, J.,
  Williams, E., Mannion, O., Parkinson, H.~E., and Brazma, A. (2014).
\newblock {Expression Atlas} update--a database of gene and transcript
  expression from microarray- and sequencing-based functional genomics
  experiments.
\newblock {\em Nucleic Acids Research\/}, {\bf 42}(Database issue), D926--32.

\bibitem[Schmidberger {\em et~al.}(2011)Schmidberger, Lennert, and
  Mansmann]{Schmidberger_et_al_2011}
Schmidberger, M., Lennert, S., and Mansmann, U. (2011).
\newblock Conceptual aspects of large meta-analyses with publicly available
  microarray data: a case study in oncology.
\newblock {\em Bioninformatics and Biology Insights\/}, {\bf 5}, 13--39.

\bibitem[Seth {\em et~al.}(2014)Seth, Shawe-Taylor, and Kaski]{Seth_et_al_2014}
Seth, S., Shawe-Taylor, J., and Kaski, S. (2014).
\newblock Retrieval of experiments by efficient comparison of marginal
  likelihoods.
\newblock In C.~Loo, K.~Yap, K.~Wong, A.~Teoh, and K.~Huang, editors, {\em
  Neural Information Processing\/}, volume 8835 of {\em Lecture Notes in
  Computer Science\/}, pages 135--142. Springer International Publishing.

\bibitem[Smith {\em et~al.}(2008)Smith, Vollrath, Bradfield, and
  Craven]{Smith_et_al_2008}
Smith, A.~A., Vollrath, A., Bradfield, C.~A., and Craven, M. (2008).
\newblock Similarity queries for temporal toxicogenomic expression profiles.
\newblock {\em PLoS Comput Biol\/}, {\bf 4}(7), e1000116.

\bibitem[Vinh {\em et~al.}(2010)Vinh, Epps, and Bailey]{Vinh_et_al_2010}
Vinh, N.~X., Epps, J., and Bailey, J. (2010).
\newblock Information theoretic measures for clusterings comparison: Variants,
  properties, normalization and correction for chance.
\newblock {\em Journal of Machine Learning Research\/}, {\bf 11}, 2837--2854.

\bibitem[Zhu {\em et~al.}(2008)Zhu, Davis, Stephens, S., and
  Chen]{Zhu_et_al_2008}
Zhu, Y., Davis, S., Stephens, R., S., M.~P., and Chen, Y. (2008).
\newblock {GEOmetadb}: powerful alternative search engine for the {Gene
  Expression Omnibus}.
\newblock {\em Bioinformatics\/}, {\bf 24}(23), 2798--2800.

\end{thebibliography}
\end{document}